\pdfoutput=1

\documentclass[11pt]{article}

\usepackage[preprint]{acl}

\usepackage{times}
\usepackage{latexsym}

\usepackage[T1]{fontenc}

\usepackage[utf8]{inputenc}

\usepackage{microtype}

\usepackage{inconsolata}

\usepackage{graphicx}

\usepackage{multirow}
\usepackage{float}

\usepackage{booktabs} 
\usepackage{multirow}
\usepackage{array}
\usepackage{caption}
\usepackage{siunitx} 
\usepackage[most]{tcolorbox}

\usepackage{tabularx}
\usepackage{makecell}
\usepackage{multirow}

\usepackage{booktabs}
\usepackage{multirow}
\usepackage[table]{xcolor}
\usepackage{array}
\usepackage{makecell}

\definecolor{familybest}{RGB}{217,226,243}
\newcommand{\revisionadd}[1]{{#1}}

\title{MuSeToM: \underline{\textbf{Mu}}lti-\underline{\textbf{Se}}nsory Evaluation of Belief Revision\\
with Textual Cues Across Six Senses}
\newcommand{\ust}{\ensuremath{^\spadesuit}}
\newcommand{\nvidia}{\ensuremath{^\clubsuit}}

\newcommand{\STAN}{\ensuremath{^\diamondsuit}}
\newcommand{\amazon}{\ensuremath{^\heartsuit}}	

\author{Chunkit Chan\ust\thanks{\quad Equal contribution.} 
\ Kawai Chung\ust$^{*}$ 
\ Sze Heng Douglas Kwok\STAN$^{*}$ 
\ Yuxuan Liu\ust 
\ \ Yauwai Yim\ust \\
\ \textbf{Tianshi Zheng\ust} 
\ \textbf{Haochen Shi\ust} 
\ \textbf{Qing Zong\ust} 
\ \textbf{Wong Kai Chung\ust} 
\ \textbf{Pak Hin Lau\ust} \\
\ \textbf{YANG Calvin\ust} 
\ \textbf{Yifan Gao\amazon}
\ \textbf{Xiaoman Pan\amazon}
\ \textbf{Ginny Wong\nvidia}\\
\ \textbf{Simon See\nvidia}
\ \textbf{Yangqiu Song\ust}\\
\ust The Hong Kong University of Science and Technology
\STAN Stanford University\\
\amazon Amazon
\nvidia NVIDIA AI Technology Center (NVAITC)\\
\texttt{\{ckchancc, yqsong\}@cse.ust.hk} \ \ \ \ 
}

\begin{document}
\maketitle
\begin{abstract}
Theory of Mind (ToM), the ability to infer mental states in others, is pivotal for human social cognition. Existing evaluations of ToM in large language models (LLMs) mainly focus on static belief attribution and false-belief reasoning, leaving a critical aspect of social reasoning underexplored: \textit{can LLMs decide whether perceptual evidence is sufficient to justify another person's belief revision?} To address this gap, we present MuSeToM, \revisionadd{a human-validated, text-based benchmark for evaluating epistemic belief revision in ToM reasoning}. MuSeToM contains 2,415 instances adapted from BigToM and OpenToM, \revisionadd{with textual cues spanning six sensory categories}, \revisionadd{and tests} whether models can determine when a protagonist should revise a belief and infer the resulting updated belief state. Using MuSeToM, we systematically evaluate state-of-the-art LLMs under direct and chain-of-thought prompting, revealing a pronounced dissonance: while models often detect that a belief may change, they struggle to identify the updated belief and frequently over-update on perceptually salient but epistemically insufficient cues\footnote{Code and data are available at \url{https://github.com/HKUST-KnowComp/MuSeToM}.}.
\end{abstract}

\section{Introduction}
\begin{figure}[!t]
    \centering
    \includegraphics[width=\columnwidth]{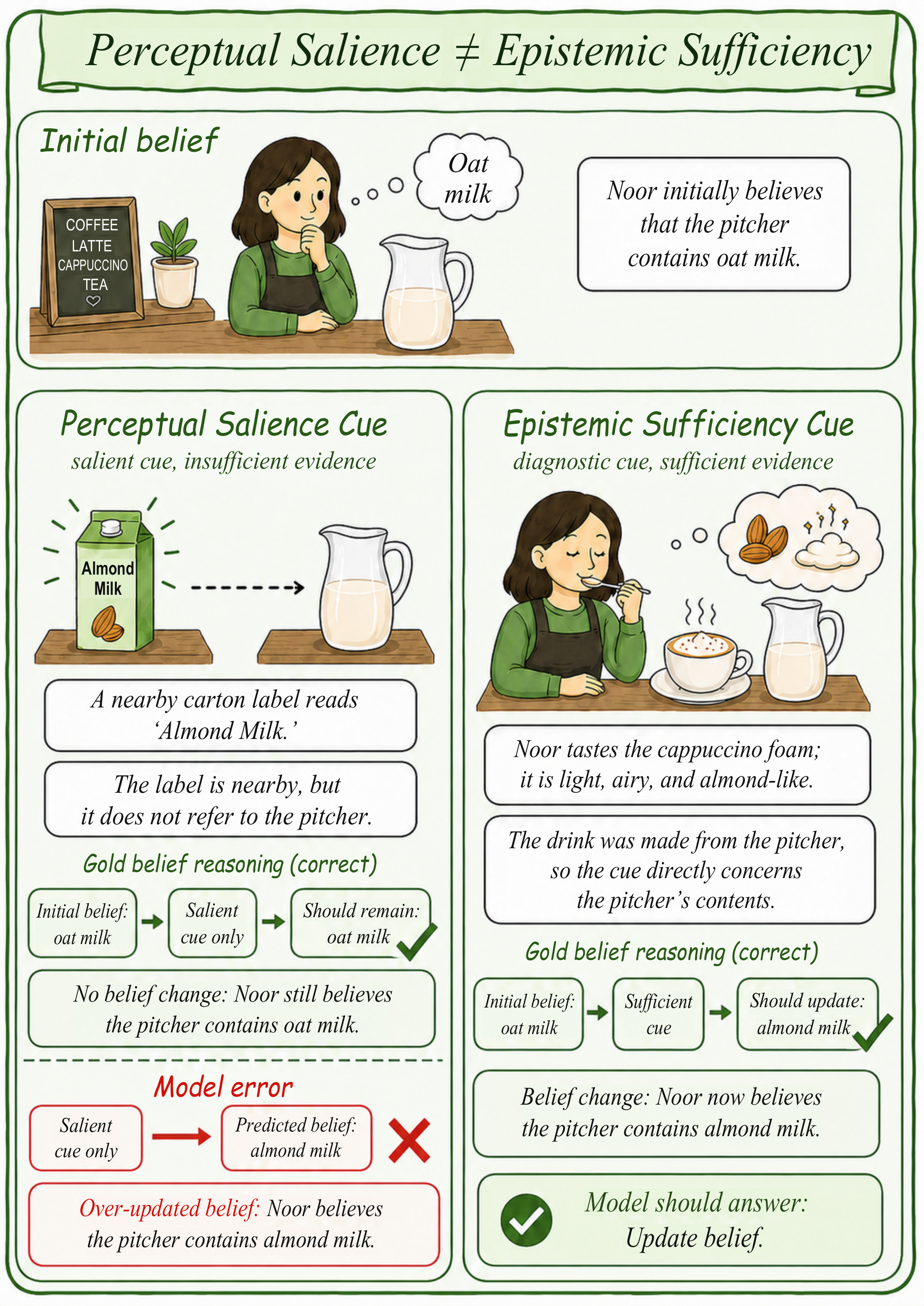}
    \caption{
    An example in MuSeToM.
    The model incorrectly treats this cue as sufficient for belief change, whereas the gold reasoning preserves Noor's prior belief.
    }
    \label{fig:intro-example}
    \vspace{-0.5cm}
\end{figure}

Theory of Mind (ToM), the ability to reason and attribute the mental states of others, is increasingly recognized as an important capability for large language models (LLMs) deployed in human-centered environments~\cite{premack1978does,ma-etal-2023-towards-holistic}. Recent studies show that modern LLMs can achieve strong performance on classical ToM benchmarks, particularly on false-belief reasoning tasks~\cite{strachan2024testing,bubeck2023sparks,kosinski2023theory}. However, most existing evaluations focus primarily on belief attribution: given a narrative context, models are asked to infer what another agent currently believes~\cite{le2019,kim2023fantom}. In contrast, a fundamental aspect of human social reasoning remains underexplored: \textit{determining whether newly observed evidence is sufficient to justify revising a prior belief.}

In realistic environments, humans continuously update their beliefs based on perceptual evidence. However, belief revision does not occur merely because new information is observed. Instead, humans evaluate whether the available evidence provides sufficient epistemic support for updating an existing belief. 
In Figure~\ref{fig:intro-example}, Noor tastes the cappuccino foam and finds it light, airy, and almond-like, providing sufficient evidence to revise her belief from oat milk to almond milk. In contrast, a nearby carton labeled ``almond milk'' may be perceptually salient and attract attention, yet such evidence alone remains insufficient to justify the same conclusion.

\textit{Perceptual salience} refers to the extent to which a stimulus stands out and captures attention due to its perceptual prominence~\cite{Itti2007VisualS,MacLeod2015ACE}.
Nevertheless, perceptually salient cues are not necessarily evidentially informative. In contrast, we use \textit{epistemic sufficiency} to describe whether available evidence provides adequate support to rationally justify belief revision under uncertainty, consistent with theories of probabilistic and causal inference in human cognition~\cite{Tenenbaum2011HowTG,Knill2004TheBB}. 
This distinction between perceptual salience and epistemic sufficiency is fundamental to human reasoning under uncertainty. Prior work in cognitive and developmental psychology further suggests that humans use perceptual evidence not only to form beliefs about the world, but also to infer when other individuals should revise their beliefs~\cite{BaronCohen1991PrecursorsTA,Gervais2013PerceivingMA, Raleigh2020PerceptualEA, Pratt1990YoungCU}. Consequently, robust ToM reasoning requires more than tracking what another agent can perceive; it also requires evaluating whether the perceived evidence is sufficiently informative to warrant belief revision.

Despite recent progress in ToM evaluation, this capability remains underexplored. 
Most prior benchmarks primarily evaluate whether models can infer another agent’s belief from a narrative context, rather than whether newly encountered perceptual evidence should trigger belief revision~\cite{le2019,kim2023fantom,xiao2025dyntom}. 
Even perception-oriented ToM work has mainly emphasized recovering perceptual access or using a character's perspective to infer current beliefs, rather than testing whether a newly encountered cue warrants revision of a prior belief~\cite{jung2024perceptions,wilf2024think}. 
Moreover, perceptual information is often simplified into coarse visibility signals (e.g., whether an event was observed), which bypasses the more challenging problem of assessing whether evidence is epistemically sufficient to support belief revision. 
As a result, current benchmarks provide limited insight into whether LLMs can distinguish genuinely informative evidence from perceptually salient but misleading or irrelevant cues.
To address this gap, we introduce MuSeToM, a benchmark for evaluating epistemic belief revision in LLM theory of mind reasoning through multi-sensory perceptual evidence. 
MuSeToM presents scenarios in which a protagonist initially holds a belief and subsequently encounters perceptual cues arising from events in the environment. These cues span multiple sensory modalities and vary in epistemic strength: some provide sufficient evidence to justify belief revision, while others are perceptually salient yet insufficient, indirect, or entirely irrelevant. \revisionadd{All model inputs are text; the six sensory categories refer to the protagonist's experiences described in the cues.}
The benchmark evaluates two core abilities: (1) determining whether the protagonist should revise their belief, and (2) identifying the resulting updated belief state. 
 
Experiments across a diverse set of state-of-the-art LLMs reveal several consistent findings. First, models perform substantially better at detecting potential belief changes than at identifying the resulting updated belief state. Second, models exhibit a systematic over-updating bias, frequently treating perceptually salient cues as sufficient evidence for belief revision even when the cues are weak or indirect. These findings suggest that current LLMs often conflate perceptual salience with epistemic sufficiency, leading to unreliable belief revision and underscoring the need for more robust and realistic evaluations of theory of mind reasoning. Our contributions are summarized as follows:
\begin{itemize}
\item We introduce MuSeToM, the first human-validated benchmark designed to evaluate epistemic belief revision in LLM theory of mind reasoning from multi-sensory perceptual evidence. MuSeToM contains 2,415 human-validated instances adapted from BigToM and OpenToM, spanning six sensory modalities and both clean and noisy reasoning settings.
\item We propose evaluation tasks that measure belief-change detection, updated belief-state identification, and epistemic relevance filtering under distractor cues.
\item Through extensive experiments across state-of-the-art LLMs, we identify systematic reasoning failures, including over-updating biases and difficulties distinguishing perceptual salience from epistemic sufficiency.
\end{itemize}

\section{Related Work}
\subsection{Theory of Mind Benchmarks}
Early computational ToM benchmarks build on classic false-belief tests such as the Sally-Anne task \citep{baron1985does}. ToM-bAbi and ToMi formulate false-belief reasoning as question answering over synthetic narratives \citep{Grant2017HowCM,Nematzadeh2018EvaluatingTO,le2019}. Subsequent work expands this setting to more realistic and challenging social reasoning scenarios. BigToM uses model-generated stories to evaluate social reasoning \citep{gandhi2023}, while FANToM stresses ToM in dialogue interactions \citep{kim2023fantom}. ToMBench provides a broader evaluation framework over social cognition abilities \citep{chen2024tombench}, and OpenToM introduces longer narratives with richer character and object descriptions \citep{xu2024opentom}. Other benchmarks study situated ToM, higher-order belief reasoning, temporal mental-state tracking, and benchmark robustness \citep{ma-etal-2023-towards-holistic,wu2023hitom,street2025higherordertom,xiao2025dyntom,ullman2023,shapira-etal-2024-clever,riemer2025brokenbenchmarks}. Recent work further extends ToM evaluation to multilingual, culturally adapted, and multimodal settings \citep{chan2025xtom,sadhu2024multitom,jin2024mmtomqa,chen2024tomseye,shi2025muma,villacueva2025moments}. These benchmarks mainly evaluate belief attribution or state tracking, which do not directly test whether newly perceived evidence is sufficient to revise another agent’s prior belief. More related works are provided in Appendix~\ref{Appendix_for_Related_Works}.

\subsection{Perception and Belief Revision}
Several recent studies examine perception and perspective-taking in ToM reasoning. PercepToM introduces Percept-ToMi and Percept-FANToM to evaluate perception inference and perception-to-belief inference \citep{jung2024perceptions}. SimToM filters the narrative context to the target character's perspective before answering ToM questions \citep{wilf2024think}. SymbolicToM tracks character belief states with explicit symbolic representations \citep{sclar-etal-2023-minding}. These works are closest to MuSeToM because they connect perceptual access, perspective context, and belief attribution. However, they focus on what an agent perceived or what belief follows from accessible context. MuSeToM instead asks whether a newly encountered perceptual cue is strong enough to revise a prior belief. Belief-R is also related because it evaluates whether language models revise their own conclusions after new evidence is introduced \citep{wilie2024belief}. In contrast, MuSeToM evaluates whether a model can infer when another agent should update a belief, and what the updated belief should be, from perceptual evidence available to that agent.
\revisionadd{Appendix~\ref{Appendix_for_Related_Works} provides a structured comparison with Belief-R.}

\begin{figure*}[t]
  \centering
  \includegraphics[width=\textwidth]{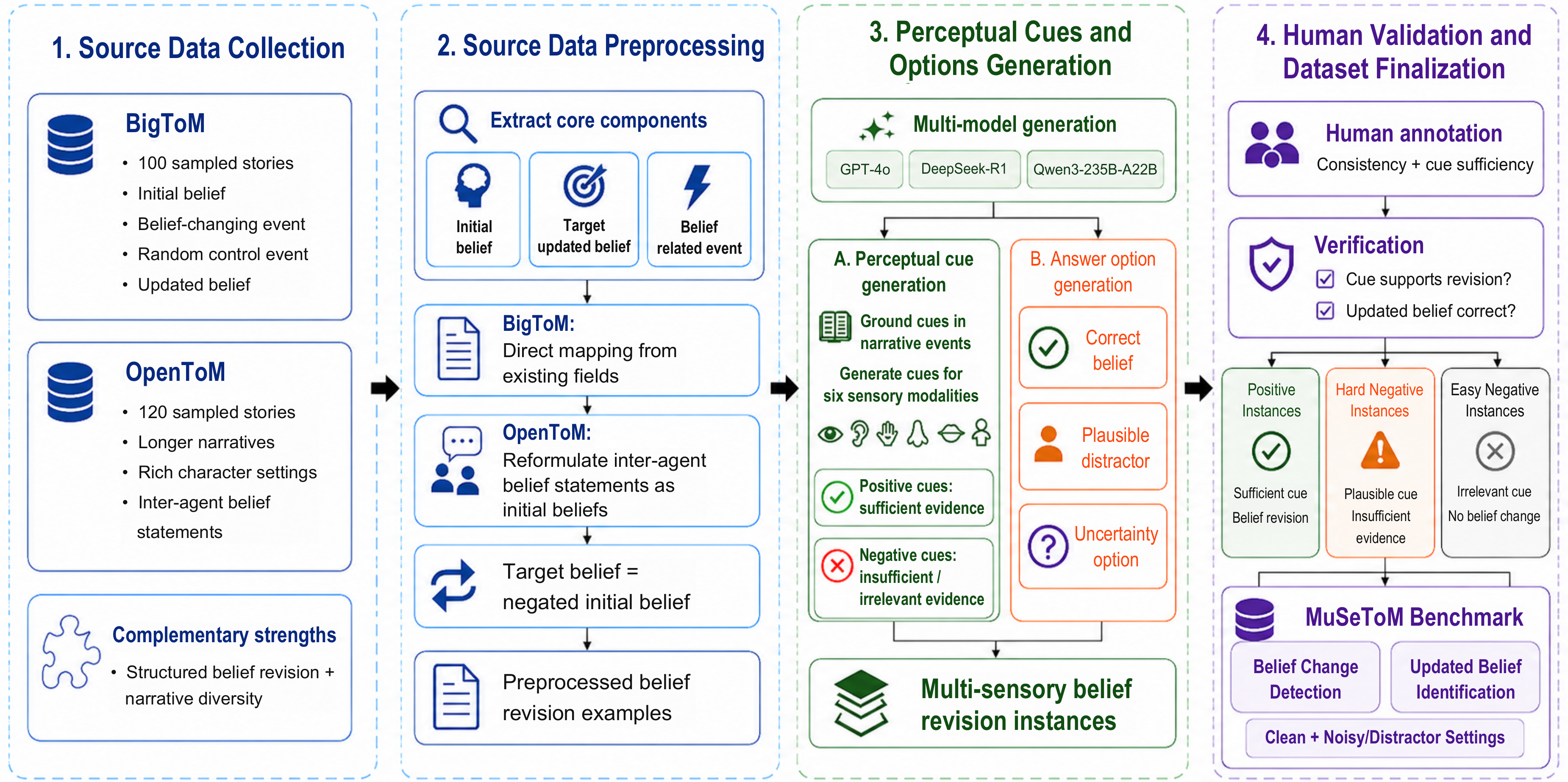}
  \caption{
  Overview of the MuSeToM dataset construction pipeline.
  }
  \label{fig:dataset-pipeline}
\end{figure*}

\section{MuSeToM}
We introduce MuSeToM, a multi-sensory benchmark designed to evaluate whether large language models (LLMs) can reason about epistemic belief revision from perceptual evidence and determine when a protagonist should revise a prior belief and infer the resulting updated belief state from perceptual cues in the environment.

\subsection{Design Considerations for MuSeToM}
A benchmark for perception-driven Theory of Mind like MuSeToM should satisfy three key requirements. First, it should provide realistic narrative contexts with sufficiently rich characters and environments to support perceptual experiences across multiple sensory modalities. Second, it should specify a clear initial belief held by the target agent. Third, it should contain, or allow the construction of, an event that either justifies revising the initial belief or confirms that the original belief should remain unchanged. These requirements are essential for evaluating epistemic belief revision rather than simple belief attribution. In particular, the benchmark must distinguish between perceptually salient observations and evidence that is genuinely sufficient to support belief change.

\subsection{Source Data Collection}
Based on these criteria, we construct MuSeToM using stories adapted from BigToM \cite{gandhi2023} and OpenToM \cite{xu2024opentom}. 
BigToM contains 200 structured stories that explicitly specify several components necessary for constructing belief-revision scenarios, including an initial belief, a belief-changing causal event, a non-belief-changing random event, and the resulting updated belief. These elements make BigToM a natural source for constructing perception-based belief revision examples.
OpenToM complements BigToM by providing 7,000 narrative scenarios with longer contexts, richer character settings, and more detailed object-centric descriptions. It also contains explicit inter-agent belief statements that can be reformulated as initial beliefs. These properties make OpenToM well suited for constructing scenarios in which perceptual evidence may or may not justify belief revision.
The two datasets provide complementary advantages: BigToM offers explicit belief revision structure, while OpenToM contributes broader narrative diversity and richer environmental detail. To assess potential data contamination risk, we report a detailed contamination check in Appendix~\ref{app:contamination}.

\subsection{Task Definition}
MuSeToM evaluates epistemic belief revision in Theory-of-Mind reasoning. \revisionadd{Given a textual story context, an initial belief, and one or more text-described sensory cues}, models are evaluated on four task formats: (1) \textit{clean belief-change detection}, which asks whether a cue provides sufficient evidence to trigger belief revision; (2) \textit{clean updated-belief identification}, which asks models to identify the resulting belief state; (3) \textit{noisy belief-change detection}, which adds a distractor cue; and (4) \textit{noisy updated-belief identification}, which tests belief-state inference under distractor cues. This design evaluates both belief revision and epistemic relevance filtering.
We evaluate categorical belief revision under the shared evidence-sufficiency criterion described in Appendix~\ref{app:normative-standard}; a no-change label does not rule out reduced confidence in the prior belief.
Detailed task questions and prompting formats are provided in Appendix~\ref{sec:Prompt_Template}.

\subsection{Dataset Construction}
\paragraph{Source Data Preprocessing}
We sample 100 stories from BigToM and 120 stories from OpenToM as the source data for MuSeToM. For each example, we extract three core components: (1) the initial belief, (2) the target updated belief, and (3) the event associated with belief revision.
For BigToM, this mapping is straightforward because the dataset already explicitly provides the relevant belief-revision components.
For OpenToM, each example involves two agents and an object embedded within a narrative context. We extract explicit inter-agent belief statements (e.g., ``[Agent 1] believes [Agent 2] likes [object]'') as the initial beliefs. The target updated belief is then constructed as the corresponding contrastive belief, while the belief-changing events are instantiated during the next stage.

\paragraph{Perceptual Cues and Options Generation}
We use multiple LLMs, including GPT-4o, DeepSeek-R1, and Qwen3-235B-A22B, to generate \revisionadd{candidate} perceptual cues grounded in the narrative events. Using multiple model families \revisionadd{aims to reduce dependence on any individual generator's linguistic style}~\citep{openai2024gpt4o,deepseek2025r1,qwen2025qwen3}.
MuSeToM covers six sensory modalities. In addition to the traditional Aristotelian five senses—visual, auditory, tactile, olfactory, and gustatory—we additionally include proprioception, which refers to awareness of body position and movement~\cite{aristotleDeAnima}. Proprioception plays an important role in how humans interact with their environment and construct beliefs about external events~\cite{Gao2019BodyIO, Biel2005RaisingAS}.
 
For each story, we generate \revisionadd{candidate cues intended to support belief revision, alongside candidates intended to be epistemically insufficient or irrelevant to the protagonist's belief. Final cue labels are assigned after human validation and filtering}. For the Updated Belief Identification task, we further generate multiple-choice answer options, including a plausible but incorrect distractor belief and an uncertainty option. The full generation prompts are provided in the Appendix, and additional details on OpenToM contrastive belief construction and noisy distractor construction are provided in Appendix~\ref{app:data-construction-clarifications}.

\paragraph{Human Validation}
All generated cues are \revisionadd{validated, filtered, and labeled} by three human annotators \revisionadd{who assess logical consistency, plausibility, and epistemic appropriateness}. Annotators determine whether each perceptual cue provides sufficient evidence to justify belief revision and verify the corresponding updated belief state. \revisionadd{Unreasonable, implausible, or logically inconsistent cues are excluded.} Based on the human validation process, the final instances are grouped into three categories:
\textbf{Positive Instances}, where the perceptual cue provides sufficient evidence for belief revision;
\textbf{Hard Negative Instances}, where the cue is \revisionadd{coherent with the story, available to the protagonist, and relevant to the target belief, but insufficient to support a definite revised belief};
\textbf{Easy Negative Instances}, where the cue is intentionally irrelevant to the protagonist’s belief.
\revisionadd{Appendix~\ref{appendix:annotation} details the normative cue-label criteria and annotation protocol.}

\subsection{Benchmark Validation}
To evaluate annotation reliability, we measure inter-annotator agreement using Fleiss’ $\kappa$ \citep{fleiss1971nominal}.
Three annotators annotated each perceptual cue with respect to two questions: whether the cue causes a belief change, and what the protagonist's updated belief is.
To improve annotation consistency, we first evaluated annotators on their initial 100 instances and provided detailed feedback on common error patterns before proceeding to the full dataset. After completing the annotation process, we filter out instances with unreasonable or implausible cues.
Despite the complexity of inferring belief revision from perceptual cues across diverse sensory modalities, we observe strong inter-annotator agreement on both source datasets. For \textit{Belief Change Detection}, Fleiss' $\kappa$ is 0.776 on BigToM and 0.742 on OpenToM. For \textit{Updated Belief Identification}, $\kappa$ is 0.744 on BigToM and 0.742 on OpenToM. 
Appendix~\ref{app:annotation-robustness} reports annotation robustness and bootstrap intervals.
Appendix~\ref{app:cue-form-probe} shows that removing the story and prior belief reduces performance in both tested cue-only models, while the cue-text probe remains above chance. Context contributes to performance, although these controls do not rule out partial reliance on cue artifacts.

\subsection{Dataset Statistics}
MuSeToM is constructed from 220 source stories, including 100 stories from BigToM and 120 stories from OpenToM. After human validation and filtering, the final benchmark contains 2,415 perceptual-cue instances spanning six sensory modalities, including 1,169 instances derived from BigToM and 1,246 derived from OpenToM. Detailed statistics are provided in Table~\ref{tab:dataset_statistics} in the Appendix.

\begin{table*}[!ht]
\centering
\small
\setlength{\tabcolsep}{4.4pt}
\renewcommand{\arraystretch}{1.12}
\begin{tabular*}{\textwidth}{@{\extracolsep{\fill}}lcccccccc@{}}
\toprule
\multirow{3}{*}{\textbf{Model / reference}}
& \multicolumn{4}{c}{\textbf{Belief Change}}
& \multicolumn{4}{c}{\textbf{Belief Change with Noise}} \\
\cmidrule(lr){2-5} \cmidrule(lr){6-9}
& \multicolumn{2}{c}{\textbf{Task 1 (Binary)}}
& \multicolumn{2}{c}{\textbf{Task 2 (Multiple Choice)}}
& \multicolumn{2}{c}{\textbf{Task 3 (Binary)}}
& \multicolumn{2}{c}{\textbf{Task 4 (Multiple Choice)}} \\
\cmidrule(lr){2-3} \cmidrule(lr){4-5} \cmidrule(lr){6-7} \cmidrule(lr){8-9}
& \textbf{Acc. (\%)} & \textbf{F1 (\%)}
& \textbf{Acc. (\%)} & \textbf{F1 (\%)}
& \textbf{Acc. (\%)} & \textbf{F1 (\%)}
& \textbf{Acc. (\%)} & \textbf{F1 (\%)} \\
\midrule

GPT-3.5-Turbo         & 44.76 & 44.55 & 41.57 & 41.60 & 64.73 & 64.51 & 62.91 & 62.79 \\
GPT-4o                & 70.23 & 70.09 & 50.02 & 49.82 & 71.73 & 71.72 & 60.01 & 59.54 \\
GPT-OSS-120B          & 82.53 & 82.53 & 61.04 & 61.08 & \textbf{80.51} & \textbf{80.48} & 70.94 & 70.88 \\
Llama-8B              & 52.09 & 51.32 & 34.20 & 34.09 & 62.37 & 60.82 & 52.26 & 51.80 \\
Llama-405B            & 77.27 & 77.25 & 54.66 & 54.63 & 71.34 & 71.13 & 59.27 & 59.15 \\
Llama-4-Maverick-17B       & 67.83 & 67.51 & 56.11 & 55.95 & 63.86 & 62.35 & 64.33 & 64.19 \\
Mistral-7B            & 55.82 & 55.26 & 41.86 & 41.79 & 60.35 & 59.67 & 58.53 & 58.61 \\
Mixtral-8x22B         & 63.52 & 63.35 & 43.60 & 43.50 & 65.88 & 65.67 & 56.71 & 56.68 \\
Qwen3-235B-A22B             & 67.83 & 67.75 & 57.10 & 57.12 & 68.71 & 68.71 & 66.76 & 66.73 \\
QwQ-32B               & 82.61 & 82.61 & 63.52 & 63.64 & 77.07 & 77.07 & 67.16 & 67.33 \\
Qwen3.5-397B-A17B     & \underline{85.34} & \underline{85.34} & \underline{74.58} & \underline{74.57} & 67.84 & 67.81 & \textbf{71.27} & \textbf{71.24} \\
DeepSeek-R1           & 77.81 & 77.54 & 70.02 & 69.97 & 68.54 & 68.24 & 67.15 & 67.10 \\
DeepSeek-V3.1         & 82.57 & 82.57 & 65.13 & 65.12 & 55.97 & 55.70 & 56.98 & 56.92 \\
DeepSeek-V3.2         & 75.65 & 75.57 & 70.72 & 70.68 & 69.86 & 69.86 & 60.62 & 60.48 \\
Kimi-K2-Thinking      & 67.41 & 65.50 & \textbf{75.40} & \textbf{75.39} & 77.07 & 77.05 & 69.45 & 69.45 \\
MiniMax-M2.1          & 78.55 & 78.54 & 73.00 & 73.00 & 76.67 & 76.67 & \underline{71.07} & \underline{71.07} \\
MiniMax-M2.5          & \textbf{86.05} & \textbf{86.05} & 74.12 & 74.10 & \underline{79.84} & \underline{79.83} & 70.73 & 70.74 \\
Phi-3-Medium-4K       & 56.56 & 55.72 & 49.77 & 49.77 & 63.79 & 63.37 & 66.42 & 66.37 \\
\midrule
Human reference & 94.29 & 94.28 & 93.83 & 93.83 & 90.76 & 90.76 & 90.02 & 89.95 \\
\midrule
\bottomrule
\end{tabular*}
\caption{Direct-prompting results on MuSeToM across clean and noisy belief-change settings. Accuracy (\%) and Macro-F1 (\%) are reported for binary belief-change detection and multiple-choice updated-belief identification. Best and second-best LLM accuracy results in each column are marked in bold and underlined, respectively.}
\label{tab:main_results_accuracy_f1}
\end{table*}

\section{Experimental Setup}
We evaluate state-of-the-art LLMs on MuSeToM under two prompting settings: direct prompting and chain-of-thought (CoT) prompting. Each model is tested on both binary belief-change detection and multiple-choice updated-belief identification.
The evaluated models cover GPT-3.5-Turbo~\citep{openai2023gpt35turbo}, GPT-4o~\citep{openai2024gpt4o}, GPT-OSS-120B~\citep{openai2025gptoss}, Llama-3.1~\citep{llama2024llama3}, Llama-4-Maverick~\citep{meta2025llama4}, Mistral-7B~\citep{jiang2023mistral7b}, Mixtral-8x22B~\citep{mistral2024mixtral8x22b}, Qwen3-235B-A22B~\citep{qwen2025qwen3}, QwQ-32B~\citep{qwen2025qwq32b}, Qwen3.5-397B-A17B~\citep{qwen2026qwen35}, DeepSeek-R1~\citep{deepseek2025r1}, DeepSeek-V3.1~\citep{deepseek2024v3,deepseek2025v31}, DeepSeek-V3.2~\citep{deepseekv32}, Kimi-K2-Thinking~\citep{kimi2025k2thinking}, MiniMax-M2.1~\citep{minimax2025m21}, MiniMax-M2.5~\citep{minimax2026m25}, and Phi-3-Medium-4K~\citep{microsoft2024phi3}.
We report results in two input settings. In the \textit{clean setting}, each instance contains a single perceptual cue. In the \textit{noisy setting}, an additional perceptual cue is included as a distractor. We use accuracy and macro F1 as aggregate metrics. To analyze belief-revision behavior, we further report true
positive rate (TPR) on positive cases, true negative rate on easy negatives (TNR-e), and true negative rate on hard negatives (TNR-h). To mitigate position bias, we randomly shuffle the order of the answer options and cue presentation for every instance~\citep{Zheng0M0H24}. We
also report a small-sample human reference based on responses from three computer science graduate students who completed the same evaluation tasks.
\revisionadd{Appendix~\ref{app:human-performance} reports human evaluation details, modality results, individual scores, and agreement.}

\section{Main Results}
Table~\ref{tab:main_results_accuracy_f1} reports direct-prompting results across four tasks. All models score below this three-evaluator reference, with especially large gaps in updated-belief identification. Clean binary detection generally yields higher accuracy than updated-belief identification: GPT-OSS-120B scores 82.53\% versus 61.04\%, and Llama-405B scores 77.27\% versus 54.66\%. Accuracy leaders also differ: MiniMax-M2.5 and Kimi-K2-Thinking lead the clean binary and multiple-choice tasks, while GPT-OSS-120B and Qwen3.5-397B-A17B lead their noisy counterparts. Clean-to-noisy binary accuracy comparisons vary by model: GPT-3.5-Turbo increases from 44.76\% to 64.73\%, whereas DeepSeek-V3.1 decreases from 82.57\% to 55.97\%. These differences motivate the following diagnostics of positive detection, hard-negative rejection, and relevance filtering.

\subsection{Models Over-Predict Belief Revision on Perceptually Salient Cues}
\label{sec:cue-calibration}

\begin{figure}[!t]
\centering
\includegraphics[width=\columnwidth]{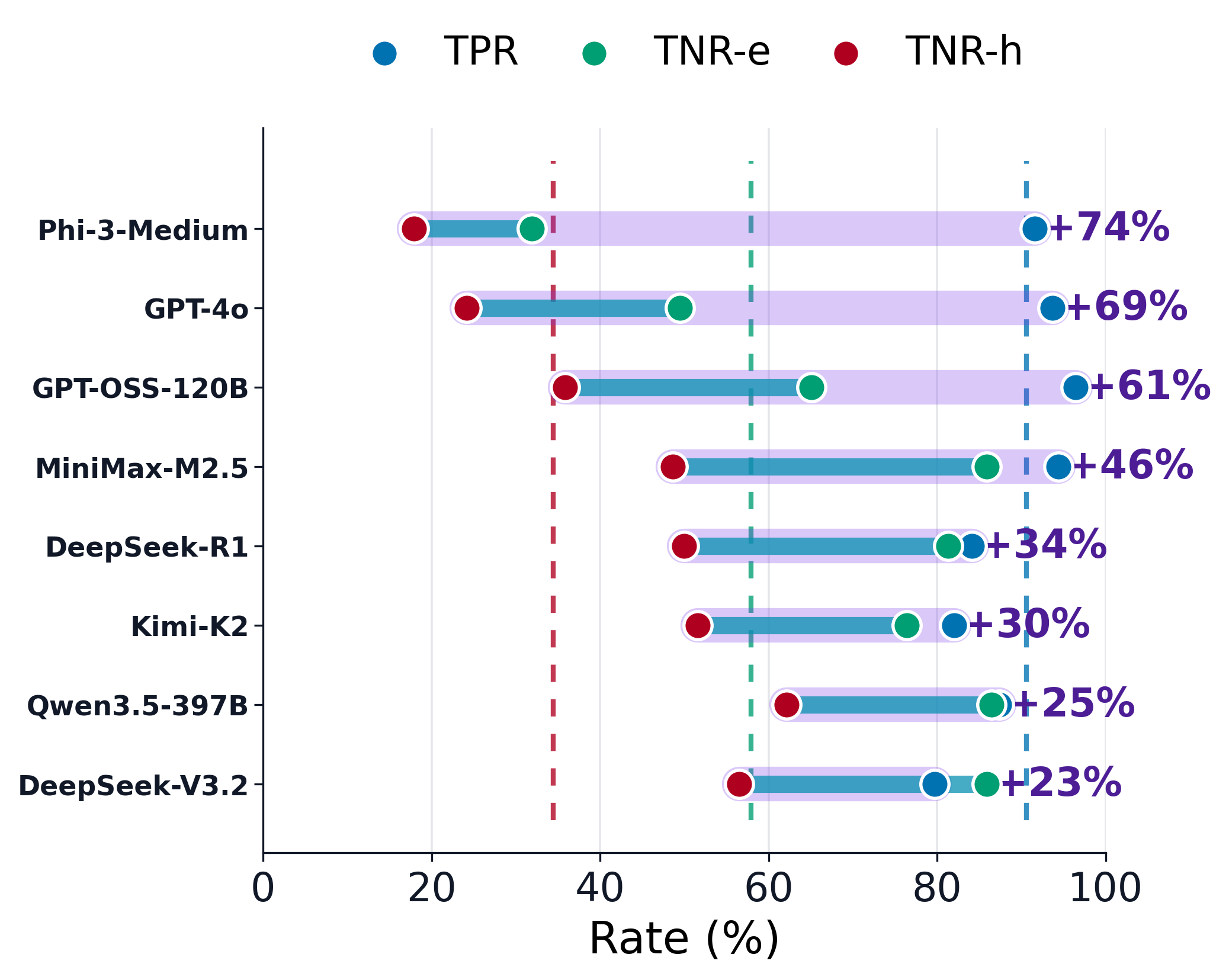}
\caption{
Diagnostic comparison of positive-cue detection and easy- and hard-negative rejection in clean direct-prompting tasks.
Dashed vertical lines mark the corresponding all-model means.
}
\label{fig:overupdating-dumbbell}
\vspace{-0.4cm}
\end{figure}

Figure~\ref{fig:overupdating-dumbbell} compares positive-cue detection with easy- and hard-negative rejection. Here, high TPR means that a model correctly predicts belief change when the cue provides sufficient evidence, while high TNR-h means that a model correctly rejects belief change when the cue is plausible but epistemically insufficient. Models generally achieve higher TPR than TNR-h, indicating better acceptance of sufficient cues than rejection of plausible but insufficient ones. The TPR--TNR-h gaps reach 74\% for Phi-3-Medium, 69\% for GPT-4o, and 61\% for GPT-OSS-120B. Even MiniMax-M2.5, the strongest clean binary model in Table~\ref{tab:main_results_accuracy_f1}, shows a 46\% gap. These gaps suggest that perceptually salient cues can be treated as sufficient evidence even when they do not justify a revised belief. The resulting asymmetry can be obscured when positive and negative cases are summarized by aggregate accuracy. The all-model means likewise show substantially higher positive-cue detection and easy-negative rejection than hard-negative rejection. Models can therefore predict no belief change, but do so less reliably for plausible, insufficient evidence than for irrelevant cues.

Appendix~\ref{app:within-context-evidence} reports a separate clean Task~1 analysis of 138 matched groups holding the story, prior belief, event, and sensory category fixed. Direct-prompting joint correctness is 26.7\% for positive--hard-negative pairs and 47.4\% for positive--easy-negative pairs. Models therefore struggle more to distinguish sufficient from relevant-but-insufficient evidence within matched contexts.

Appendix~\ref{app:hn-semantic-errors} reports a separate clean Task~2 analysis: under direct prompting, 49.4\% of hard-negative responses adopt the target belief and 25.9\% attribute uncertainty to the protagonist. Non-prior responses therefore include both unsupported belief replacement and attributed uncertainty. These percentages describe response categories, not the models' internal confidence; the uncertainty option concerns the protagonist's belief state. This distinction matters when interpreting an incorrect response as an unsupported categorical belief change.

\subsection{Distractor Cues and CoT Test Epistemic Relevance Filtering}
\label{sec:noise-cot}

\begin{figure*}[t]
    \centering
    \includegraphics[width=\textwidth]{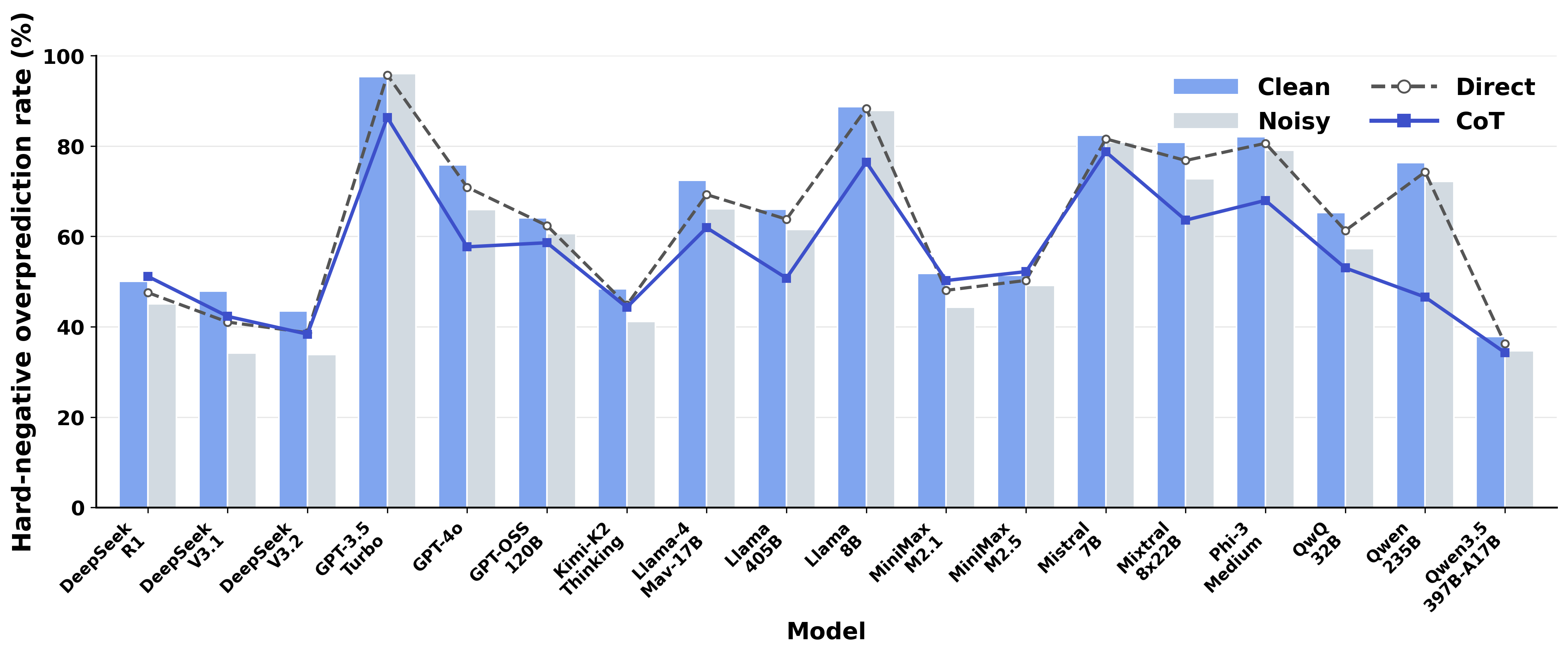}
    \caption{
    Hard-negative overprediction rates across clean and noisy settings under direct and chain-of-thought prompting.
    Bars report setting-level hard-negative overprediction rates, defined as $1-\mathrm{TNR}\mbox{-}h$, while lines compare direct and chain-of-thought prompting.
    }
    \label{fig:noise-overprediction}
    \vspace{-0.4cm}
\end{figure*}

Figure~\ref{fig:noise-overprediction} compares hard-negative overprediction across clean and noisy settings, while Table~\ref{tab:cot-filtering-gain} decomposes CoT effects into changes in positive-cue detection and hard-negative rejection. Distractor cues require models to identify relevant evidence before deciding whether belief revision is warranted, making noisy settings a stress test for epistemic relevance filtering. In particular, aggregate performance alone may obscure whether a model succeeds by detecting sufficient belief-changing cues, by rejecting hard negatives, or by shifting its response bias between these two behaviors. A lower change rate can improve rejection while reducing positive detection, so both components are needed to interpret the effect of CoT.

\begin{table}[!t]
\centering
\small
\begin{tabular}{lrrr}
\toprule
Model & $\Delta\mathrm{TPR}$ & $\Delta\mathrm{TNR}\mbox{-}h$ & Filtering gain \\
\midrule
Phi-3-Medium   & -13.7 & 10.2 & 23.9 \\
GPT-4o         & 2.3 & 9.8 & 7.5 \\
GPT-OSS-120B   & -0.6 & 3.8 & 4.4 \\
MiniMax-M2.5   & -2.4 & -1.5 & 1.0 \\
DeepSeek-R1    & 9.1 & -3.1 & -12.2 \\
Kimi-K2-Thinking & -12.8 & -1.7 & 11.1 \\
Qwen3.5-397B   & 1.2 & 3.6 & 2.3 \\
DeepSeek-V3.2  & 0.4 & -1.7 & -2.1 \\
\midrule
Mean           & -2.1 & 2.4 & 4.5 \\
\bottomrule
\end{tabular}
\caption{
Effect of chain-of-thought prompting on epistemic relevance filtering for selected models under noisy settings.
Filtering gain is defined as $\Delta\mathrm{TNR}\mbox{-}h - \Delta\mathrm{TPR}$.
}
\label{tab:cot-filtering-gain}
\vspace{-0.4cm}
\end{table}
Figure~\ref{fig:noise-overprediction} shows that hard-negative overprediction remains high across both clean and noisy conditions, indicating that distractor cues do not eliminate the tendency to over-predict belief revision on plausible but insufficient evidence. Several models, including GPT-3.5-Turbo, Llama-8B, Mistral-7B, Mixtral-8x22B, Phi-3-Medium, and Qwen3-235B-A22B, show hard-negative overprediction rates around or above 70\% in at least one direct-prompting setting. At the same time, noisy inputs do not produce a uniform degradation pattern. For some models, such as DeepSeek-V3.2, noisy hard-negative overprediction is lower than clean hard-negative overprediction, whereas other models continue to show high overprediction under noise. This suggests that noisy settings change the balance between accepting sufficient evidence and rejecting plausible but insufficient cues, rather than simply making all examples harder. Table~\ref{tab:cot-filtering-gain} further shows why aggregate performance can mask different effects on positive detection and hard-negative rejection. If CoT mainly improves epistemic relevance filtering, gains in $\mathrm{TNR}\mbox{-}h$ should exceed gains in $\mathrm{TPR}$, meaning that the model becomes better at rejecting hard negatives rather than merely predicting more belief changes. On average, CoT slightly decreases positive-cue detection by 2.1\% while increasing hard-negative rejection by 2.4\%, producing a mean filtering gain of 4.5\%. However, this average hides substantial model-level variation. Phi-3-Medium shows the largest filtering gain, 23.9\%, because CoT reduces TPR by 13.7\% while increasing TNR-h by 10.2\%. GPT-4o and GPT-OSS-120B show more direct filtering improvements, with TNR-h increasing by 9.8\% and 3.8\%, respectively, while TPR changes only modestly. In contrast, DeepSeek-R1 shows a negative filtering gain of -12.2\%, because TPR increases by 9.1\% while TNR-h decreases by 3.1\%. DeepSeek-V3.2 also shows a negative filtering gain, indicating that CoT does not consistently improve hard-negative rejection. These results show that noisy evaluation can reveal different trade-offs between positive detection and hard-negative rejection that are obscured by aggregate accuracy or F1.

\subsection{Aggregate Accuracy Masks Modality-Level Hard-Negative Rejection Failures}
\label{sec:modality-calibration}

\begin{figure}[!t]
\centering
\includegraphics[width=0.9\linewidth]{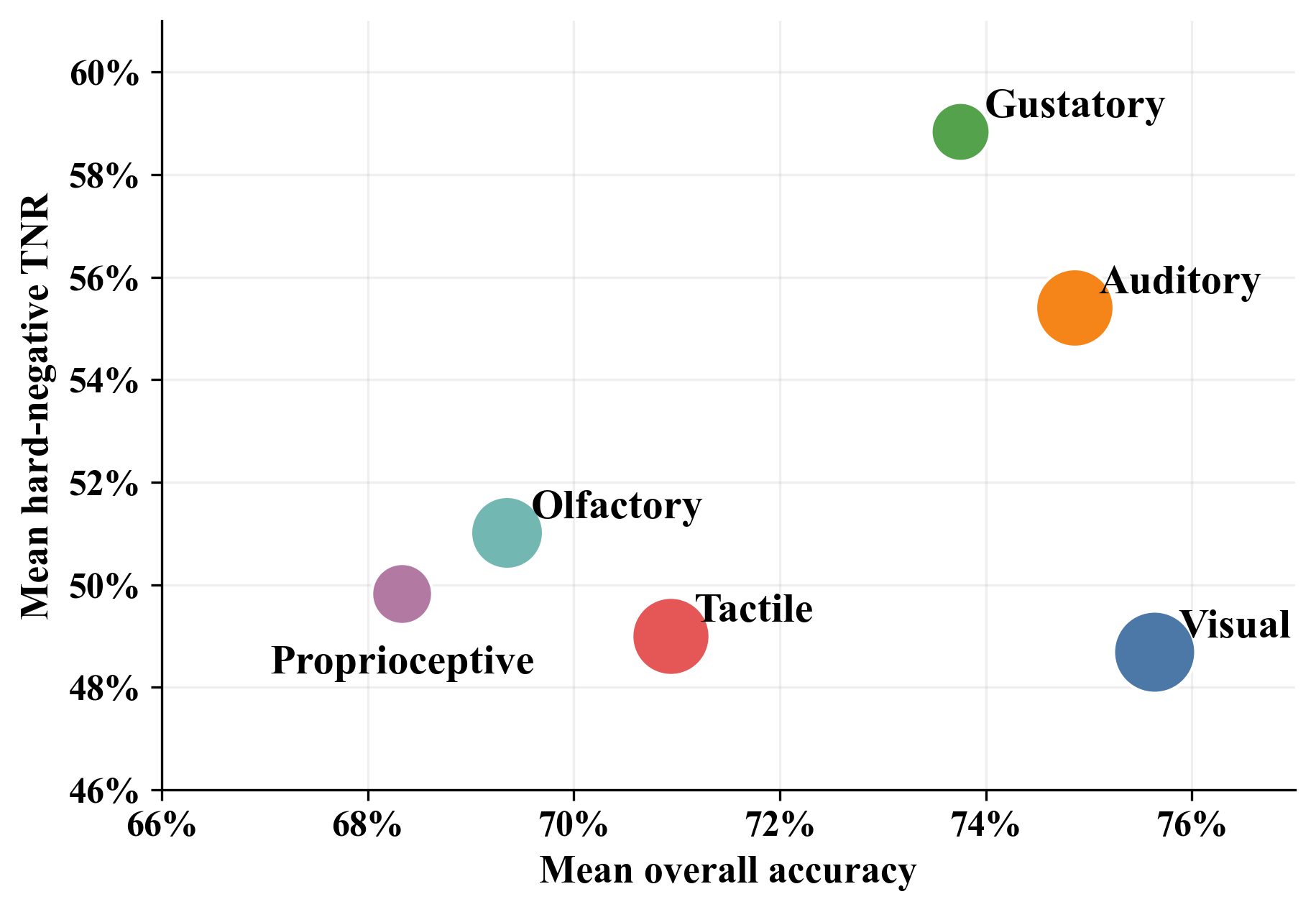}
\caption{
Modality-level comparison between overall accuracy and hard-negative rejection (TNR-h).
}

\label{fig:modality-tradeoff}
\vspace{-0.4cm}
\end{figure}

Figure~\ref{fig:modality-tradeoff} compares macro-averaged overall accuracy with macro-averaged TNR-h across sensory modalities. Visual cases achieve the highest mean accuracy of 75.6\% but the lowest TNR-h of 48.7\%, whereas gustatory cases have lower overall accuracy but the strongest TNR-h of 58.8\%. Thus, aggregate accuracy can mask failures to reject plausible but insufficient evidence within a modality.
\revisionadd{These comparisons characterize reasoning over text-described sensory categories and do not isolate intrinsic sensory capabilities.}
Appendix~\ref{app:modality-composition-controls} shows that category-standardized clean Task~1 accuracy ranks gustatory above visual cues and that hard-negative rejection varies by source. Modality rankings therefore depend partly on item composition. Category standardization holds the proportions of positive, hard-negative, and easy-negative cases constant; it does not equalize source composition. Source-stratified rejection rates therefore complement the standardized comparison.

\section{Conclusion}
MuSeToM evaluates LLM Theory-of-Mind belief revision from text-described sensory cues through belief-change detection and updated-belief identification, in both clean and noisy settings. Models often accept sufficient evidence yet systematically over-predict revision when perceptually salient cues are plausible but epistemically insufficient. Noisy evaluation and modality-level diagnostics further show how aggregate performance can obscure failures to reject insufficient evidence. These findings highlight the need to distinguish perceptual salience from evidential sufficiency when evaluating belief revision in LLMs.

\clearpage
\section*{Limitations}

\paragraph{Generated Perceptual Cues}
\revisionadd{LLM-generated cues may differ stylistically from naturally occurring perceptual descriptions. Using multiple generator families and human validation aims to mitigate these artifacts: annotators assess logical consistency, plausibility, and epistemic appropriateness and remove unsuitable cues. These measures do not establish that generation artifacts are absent. The cue-text probes and cue-only LLM evaluations in Appendix~\ref{app:cue-form-probe} provide additional diagnostics, while leaving possible partial reliance on cue artifacts unresolved.}

\paragraph{Potential Contamination}
Most existing NLP benchmarks were released before the training of many recent LLMs, meaning that some benchmark data may have been seen during pre-training or post-training~\citep{golchin2024time,li2024task}. Since MuSeToM is constructed from BigToM and OpenToM, potential data contamination cannot be completely ruled out. However, MuSeToM substantially transforms the original benchmark instances into new multi-sensory belief-revision examples through generated perceptual cues, noisy distractors, updated-belief options, and human validation. We further conduct free-generation and partial-reconstruction contamination checks on sampled source instances. Across the evaluated models, none of the generated responses matched the sampled BigToM or OpenToM instances under our matching criteria. These results suggest that the sampled source stories used in MuSeToM are not identified as suffering from direct data contamination. Therefore, although contamination remains a possible limitation, it may not be a primary concern for MuSeToM. Details are provided in Appendix~\ref{app:contamination}.

\section*{Ethics Statement}

In this work, we conformed to recognized privacy practices and rigorously followed the data usage policy. This paper introduces MuSeToM, a multi-sensory benchmark for assessing theory of mind belief revision in large language models. MuSeToM is constructed from existing theory of mind benchmarks, including BigToM and OpenToM, and further transforms their original stories into multi-sensory belief-revision instances through newly generated perceptual cues, answer options, noisy distractors, and human validation. We conduct human validation to ensure the logical consistency, plausibility, and epistemic appropriateness of the generated perceptual cues. In particular, annotators verify whether each cue provides sufficient evidence to trigger belief revision and whether the corresponding updated belief state is valid. We also filter out unreasonable, implausible, or potentially inappropriate instances during the validation process. We do not anticipate immediate harmful social consequences from the benchmark itself; however, because the benchmark is derived from existing datasets and generated cues, future work should audit MuSeToM for possible social or cultural bias and for artifacts introduced by cue generation or filtering. Moreover, the licenses and usage conditions of the source datasets allow us to use and adapt them for evaluation and research purposes. We follow the intended research use of these resources and will clearly acknowledge the original datasets.

\paragraph{Resource Copyright}
This work presents a new resource: MuSeToM, which is a multi-sensory extension of BigToM and OpenToM for evaluating perception-driven belief revision in theory of mind reasoning. Both source datasets are publicly available for research purposes, and we do not add any additional restrictions for accessing MuSeToM beyond the terms required by the original resources. We will highlight the sources of our data and ask users to cite the original BigToM and OpenToM papers when they use MuSeToM for research.

\section*{Acknowledgements}
The authors of this paper were supported by the ITSP Platform Research Project (ITS/189/23FP) from ITC of Hong Kong, SAR, China, and the AoE (AoE/E-601/24-N), the RIF (R6021-20) and the GRF (16205322) from RGC of Hong Kong, SAR, China. We also thank the support from NVIDIA AI Technology Center (NVAITC) and Amazon.

\bibliography{custom}
\clearpage

\appendix

\section{Appendix for Data Construction and Validation}

\label{app:data-validation}

The appendix text is organized in Appendices~A--F. All supporting tables and figures are collected after the text in Appendix~\ref{app:supplementary-tables-figures}. 
Appendix~\ref{app:data-construction} describes cue construction and sensory categories, while Appendix~\ref{appendix:annotation} covers annotation and disagreement. Appendix~\ref{app:dataset-annotation-tables} reports dataset statistics; Appendix~\ref{app:cue-form-probe} presents cue-artifact diagnostics; and Appendix~\ref{app:contamination} checks source-story contamination.

\subsection{Appendix for Data Construction}
\label{app:data-construction}
\paragraph{Data Construction}
We construct MuSeToM by augmenting BigToM and OpenToM stories with perceptual cues and events. For BigToM, the generation prompts produce perceptual cues from the original causal-event and random-event fields. For OpenToM, the generation prompts produce belief-changing one-line events and unrelated random one-line events. The generated candidates cover six sensory modalities: visual, auditory, olfactory, gustatory, tactile, and proprioception. By generating perceptual cues from both belief-relevant and unrelated events, we obtain a balanced pool of candidate instances that vary in their relevance to the protagonist's initial belief. After human annotation and filtering, retained instances are labeled as positive, hard-negative, or easy-negative cues. Table~\ref{tab:dataset_statistics} reports the final number of source stories, validated cue instances, and modality-specific cue counts after human filtering.

We note that BigToM random-event cue generation uses the same story--belief--event prompt structure as the causal-event setting, but replaces the causal event with the original BigToM random-event field and generates one cue per modality. Since the substituted random event is unrelated to the belief-changing event, the resulting cues are intended to be irrelevant to the belief-changing event, with final labels determined after human annotation and filtering.

\paragraph{Generation Prompt}
The generation prompt templates used for BigToM and OpenToM are shown in Tables~\ref{tab:bigtom-generation-prompt} and~\ref{tab:opentom-generation-prompt}. These prompts are used only for generating candidate cues and events; all retained instances are subsequently validated by human annotators.

\subsubsection{Source-Specific and Noisy Instance Construction}
\label{app:data-construction-clarifications}

\paragraph{OpenToM contrastive belief construction.}
OpenToM instances provide narrative contexts with two agents, an object, and inter-agent beliefs about object preferences. In MuSeToM, we use these cases as controlled contrastive belief-revision settings. The ``negation'' used during generation should therefore be understood as a target contrastive belief for preference-style beliefs, not as a claim that belief revision is always simple logical negation. Candidate OpenToM events are retained only when the resulting updated belief is natural in the story, coherent with the observed character's preference, and supported by the perceptual cue. During human validation, annotators verify both whether the cue warrants belief revision and whether the updated belief option is appropriate. Cases that require uncertainty, partial revision, or an unnatural contrastive belief are filtered out.

\paragraph{Noisy distractor construction.}
The noisy setting is designed to test epistemic relevance filtering rather than introduce arbitrary out-of-context noise. For each noisy example, a task-relevant cue from the validated positive or hard-negative pool is paired with an additional distractor cue sampled from the same story's easy-negative pool. These distractor cues are coherent with the story context but do not provide sufficient evidence for the queried belief revision. Distractor position is randomized across examples so that models cannot rely on cue order. The same answer-label randomization used in the clean setting is preserved in the noisy setting. As the added distractor is chosen from the easy-negative pool, it does not determine the answer by itself. This construction makes the noisy setting a test of whether a model can identify which perceptual evidence is relevant and whether that evidence is epistemically sufficient.

\subsubsection{Sensory Category Definitions}\label{app:sensory-definitions}
Table~\ref{tab:sensory_system_definitions} summarizes the six sensory systems used in MuSeToM, adapted from the STAR Institute's broader eight-sense framework~\citep{star2026eightSenses}.

\subsection{Appendix for Human Annotation}
\label{appendix:annotation}
We employ three human annotators to validate the generated MuSeToM instances. Annotators judge each perceptual cue with respect to two questions: whether the cue provides sufficient evidence for the protagonist to revise the initial belief, and what the protagonist's updated belief should be when belief revision is warranted. Before the full annotation round, annotators complete an initial calibration batch of 100 instances. We inspect disagreements in this batch and provide feedback on common error types, especially cases where a cue is perceptually salient but not evidentially sufficient.

\paragraph{Operational criteria for cue labels.}
\revisionadd{These labels provide a normative reference for whether the available evidence justifies definite belief revision under the stated context; they do not describe every individual's response to weak evidence. Table~\ref{tab:cue-operational-criteria} summarizes the operational criteria. Annotator disagreement triggers adjudication when needed, but is neither necessary nor sufficient for a hard-negative label. Both initially unanimous no-change cases and disputed cases adjudicated as insufficient are retained when they meet the same criteria. The annotation and adjudication procedure is described below.}

A positive cue supports the annotated revised belief. A hard negative bears on the target belief but remains indirect, ambiguous, or compatible with multiple explanations, so it does not support a definite revision. An easy negative does not bear on the target belief. Incoherent or implausible cues are excluded. Annotator disagreement is not a labeling criterion: retained hard negatives include both unanimous judgments and adjudicated cases, as reported in Appendix~\ref{app:annotation-robustness}. \revisionadd{The hard-negative label concerns the insufficiency of the evidence, even when that insufficiency is unanimously judged.} These rules operationalize evidential support without specifying a numerical probability threshold or requiring deductive entailment.

\paragraph{Rationale and scope of the shared standard.}
\label{app:normative-standard}
MuSeToM uses a shared sufficiency criterion to make categorical belief-revision judgments comparable across models. For a fixed story and cue, the evaluation asks whether the evidence available to the protagonist warrants the specified revised belief. Allowing unspecified personal thresholds would leave the reference answer dependent on unobserved factors such as prior confidence or risk preferences. Human validation and adjudication therefore establish a common operational reference for distinguishing warranted revision from responses to merely salient or relevant cues. Agreement under this protocol measures consistency in applying the reference; it does not establish a uniquely correct standard for human belief revision.

The tradeoff is that the benchmark does not characterize the full distribution of human updating behavior. For hard negatives, ``no change'' means that the cue does not sufficiently support categorical replacement of the stated belief; it does not rule out reduced confidence in the retained belief. Modeling heterogeneous thresholds would address the complementary objective of predicting individual responses and would require additional information about those thresholds or a reference distribution of judgments. MuSeToM scores should therefore be interpreted as performance relative to the stated evidence-based annotation criteria, rather than as predictions of how every individual would respond.

\revisionadd{During final annotation, annotators apply the same operational criteria to all cues. Initially unanimous no-change cues that remain plausible and belief-relevant but insufficient are retained as hard negatives. Cases without an initial consensus are discussed and adjudicated using the same criteria; a disputed cue is retained as a hard negative when the annotators agree that it is plausible and relevant but insufficient for justified belief revision. Disagreement serves as a quality-control signal for adjudication and does not define the category. Unreasonable, implausible, or logically inconsistent cues, and cases that remain unresolved after discussion, are excluded.} Table~\ref{tab:kappa} reports Fleiss' $\kappa$ for both annotation questions on BigToM and OpenToM.
Figure~\ref{fig:annotation-interface} shows the instruction and the interface provided to annotators.
\subsubsection{Annotation Disagreement and Sampling Uncertainty}
\label{app:annotation-robustness}

We examine whether hard-negative errors are concentrated on cases that annotators initially disputed. Among the 477 retained hard negatives, 330, or 69.2\%, received unanimous no-change judgments, while 147, or 30.8\%, received one change vote out of three before adjudication. We compare model predictions across these two annotation strata.

\paragraph{Evaluation and aggregation.}
We use clean binary belief-change detection (Task~1) under direct and CoT prompting. For each item, we average its valid repetitions, then average items within each model--prompt condition, and finally give the 18 models $\times$ two prompting conditions equal weight. Invalid responses are excluded from the denominators. This Task~1-only, item-weighted analysis is distinct from the clean direct-prompting diagnostic in Figure~\ref{fig:overupdating-dumbbell}, which pools valid answer slots across Tasks~1 and~2.

\revisionadd{We estimate a percentile 95\% confidence interval for the disputed-minus-unanimous gap using 10,000 bootstrap resamples of source-story groups. We first average each item's error rate across model--prompt conditions, then retain all hard-negative cues belonging to each sampled story together. This interval quantifies story-sampling uncertainty for the fixed evaluated model cohort.}

\revisionadd{Table~\ref{tab:annotation-robustness} shows a higher change-prediction rate of 58.9\% on disputed hard negatives versus 43.3\% on unanimous hard negatives, a difference of 15.6 percentage points, with a story-clustered 95\% CI of 10.7--20.6. The difference has the same direction in 32 of 36 model--prompt conditions. Initial disagreement is associated with higher difficulty, but substantial over-updating remains on the unanimously labeled subset. This comparison does not establish a universal human threshold for belief revision.}

\paragraph{Qualitative analysis of disputed cases.}
\label{app:disputed-case-analysis}
We examine three retained hard negatives with recorded binary judgments of one change vote and two no-change votes, and final gold labels of no change. The cases illustrate different gaps between a salient cue and the specific revised belief. The available records preserve votes and final labels, but not written adjudication rationales; the competing readings below are retrospective applications of the annotation criteria, not accounts of individual annotators' reasoning.

\paragraph{Linking a cue to its referent (BigToM, 0-0-1; visual).}
Noor believes that a pitcher contains oat milk. The cue describes an ``Almond Milk'' label on a nearby carton. One reading links the carton to the pitcher's contents and favors changing the belief to almond milk. Another distinguishes the observed carton from the pitcher: the cue does not establish that milk from this carton entered the pitcher. Under the annotation criteria, the label is relevant but does not sufficiently support the specific replacement belief. The final gold answer retains the oat-milk belief.

\paragraph{Inferring a cause from an effect (BigToM, 2-0-2; olfactory).}
Carlos believes that a water valve is closed. He smells decaying plant matter and rain-soaked earth from a flooded field. Flooding may suggest an open valve, but the described odor does not distinguish water released through that valve from water accumulating through rainfall or another route. The cue therefore bears on the irrigation situation without identifying the valve's state. The final gold answer retains the closed-valve belief.

\paragraph{Separating past and current preferences (OpenToM, 1-0-4; auditory).}
Paxton believes that Hadley dislikes a hat. As Hadley is about to throw it away, Paxton hears Hadley say fondly, ``This used to be my favorite.'' The affectionate tone may suggest liking, whereas ``used to'' locates the preference in the past and is compatible with present dislike. The cue does not establish that Hadley currently likes the hat. The final gold answer retains Paxton's belief that Hadley dislikes it.

These cases illustrate how relevance, causal plausibility, or affective salience can invite a change judgment without sufficiently supporting the proposed revised belief. They show possible sources of disagreement under the same operational criteria, without estimating how frequently each occurs across the disputed subset.

\subsection{Appendix for Dataset Statistics and Annotation Agreement}
\label{app:dataset-annotation-tables}

Table~\ref{tab:dataset_statistics} summarizes the number of validated MuSeToM instances by source dataset, belief-change category, and sensory modality. Table~\ref{tab:kappa} reports inter-annotator agreement for the two validation questions used to decide belief-change labels and updated belief states.

\subsection{Appendix for Cue-Text and Cue-Only Validation}
\label{app:cue-form-probe}

We assess label predictability without the surrounding story using all 1,489 positive and hard-negative cues, comprising 1,012 positive and 477 hard-negative cues grouped into 220 distinct story-text groups. We use five-fold stratified cross-validation grouped by story: all cues from a story remain in the same fold, and preprocessing is fitted on training folds only. A metadata probe uses source dataset, sensory category, word count, and character count. A cue-text probe combines word unigram/bigram and character 3--5-gram TF--IDF features. Both use class-balanced logistic regression and receive no story context, answer options, or gold-answer-derived features. Metrics are computed from pooled out-of-fold predictions.

Table~\ref{tab:cue-form-probe} shows that the cue-text probe achieves 67.8\% balanced accuracy, compared with 60.7\% for metadata and length. This result indicates that cue wording carries label information, but it does not distinguish unintended generation artifacts from the intended semantic difference in evidential strength. Nor does it establish whether the evaluated LLMs exploit these features. The probe therefore characterizes cue-text predictability without ruling out artifact-based shortcuts.

\paragraph{Cue-only LLM evaluation.}
We additionally evaluate MiniMax-M2.5 and Kimi-K2-Thinking with the story context and the protagonist's initial belief removed.

Table~\ref{tab:cue-only-llm} shows lower cue-only performance for both models. Accuracy drops by 30.12 and 37.05 percentage points on Tasks 1 and 2 for MiniMax-M2.5, and by 19.65 and 37.69 points for Kimi-K2-Thinking. Thus, cue-only inputs do not recover full-context performance for either model. This complements the TF--IDF probe: cue wording carries label information, while the removed context contributes substantially to LLM performance. These results do not rule out partial reliance on cue artifacts.

\subsection{Appendix for Verification of Potential Contamination}
\label{app:contamination}

Since MuSeToM is constructed from existing ToM benchmarks, namely BigToM and OpenToM, it may be exposed to potential data contamination risks. These source datasets were released before the training of many recent LLMs and therefore could have appeared during either pre-training or post-training, including supervised fine-tuning~\citep{ouyang2022training} or RLHF~\citep{christiano2017deep}.
To assess this risk, we follow two established contamination-detection protocols. First, following Li and Flanigan~\citep{li2024task}, we use a free-generation check that asks a model to generate an example and answer from a named source benchmark. Second, following Golchin and Surdeanu~\citep{golchin2024time}, we use a partial-reconstruction check that asks a model to reconstruct the next sentence from the first sentence of a sampled source instance.
For each source dataset, evaluated model, and protocol, we run 100 trials and compare model outputs against the corresponding ground-truth source instances. All responses are screened by automatic matching and then manually reviewed under the strong, likely, and possible match criteria reported in Table~\ref{tab:contamination_check}. The prompt templates are shown in Table~\ref{tab:contamination_prompt_template}.
Across BigToM and OpenToM, none of the evaluated LLMs' generated responses match the sampled source instances under either protocol. This suggests that the sampled source stories used in MuSeToM are not identified as suffering from direct data contamination. Moreover, MuSeToM transforms the original stories into multi-sensory belief-revision instances through newly generated perceptual cues, noisy distractors, updated-belief options, and human validation, which further reduces the likelihood that models can solve the benchmark by memorizing the original datasets alone. Therefore, although data contamination cannot be completely ruled out, it may not be a primary concern for MuSeToM.

\section{Appendix for Evaluation Setup and Human Reference}
\subsection{Appendix for Hyperparameters}
\label{sec:hyparameter}

We use the following hyperparameters for assessing the large language models mentioned in this paper. For all Task 1--4 evaluations on MuSeToM, we use \texttt{temperature}=0 and \texttt{top\_p}=1.0. For Llama-3.1-8B-Instruct~\citep{llama2024llama3}, Llama-3.1-405B-Instruct~\citep{llama2024llama3}, Llama-4-Maverick-17B-128E-Instruct~\citep{meta2025llama4}, Mistral-7B-Instruct-v0.3~\citep{mistral2024instructv03}, Mixtral-8x22B-Instruct-v0.1~\citep{mistral2024mixtral8x22b}, QwQ-32B~\citep{qwen2025qwq32b}, Qwen3-235B-A22B~\citep{qwen2025qwen3}, GPT-OSS-120B~\citep{openai2025gptoss}, GPT-4o~\citep{openai2024gpt4o}, and GPT-3.5-Turbo~\citep{openai2023gpt35turbo}, we set \texttt{max\_tokens}=1024. For DeepSeek-V3.1~\citep{deepseek2024v3,deepseek2025v31}, DeepSeek-V3.2~\citep{deepseekv32}, DeepSeek-R1~\citep{deepseek2025r1}, MiniMax-M2.1~\citep{minimax2025m21}, MiniMax-M2.5~\citep{minimax2026m25}, Kimi-K2-Thinking~\citep{kimi2025k2thinking}, and Qwen3.5-397B-A17B~\citep{qwen2026qwen35}, we set \texttt{max\_tokens}=10240.

\subsection{Appendix for Evaluation Prompting}
\label{sec:Prompt_Template}
Appendix~\ref{app:evaluation-prompts} indexes the full evaluation prompt templates.
In this study, there are two types of prompting methods. The first one is zero-shot prompting, and we use the prompt templates shown in Tables~\ref{tab:musetom-task1-prompt}--\ref{tab:musetom-task4-prompt}. The second one is the chain-of-thought (CoT) prompting method, which follows~\citet{DBLP:conf/nips/Wei0SBIXCLZ22} and uses the prompt ``Let's think step by step.'' The CoT prompting template is formed by appending ``Let's think step by step.'' to the zero-shot prompt template.

\subsection{Appendix for Human Reference Evaluation}
\label{app:human-performance}

\begin{figure*}[!htbp]
\centering
\includegraphics[width=0.82\textwidth]{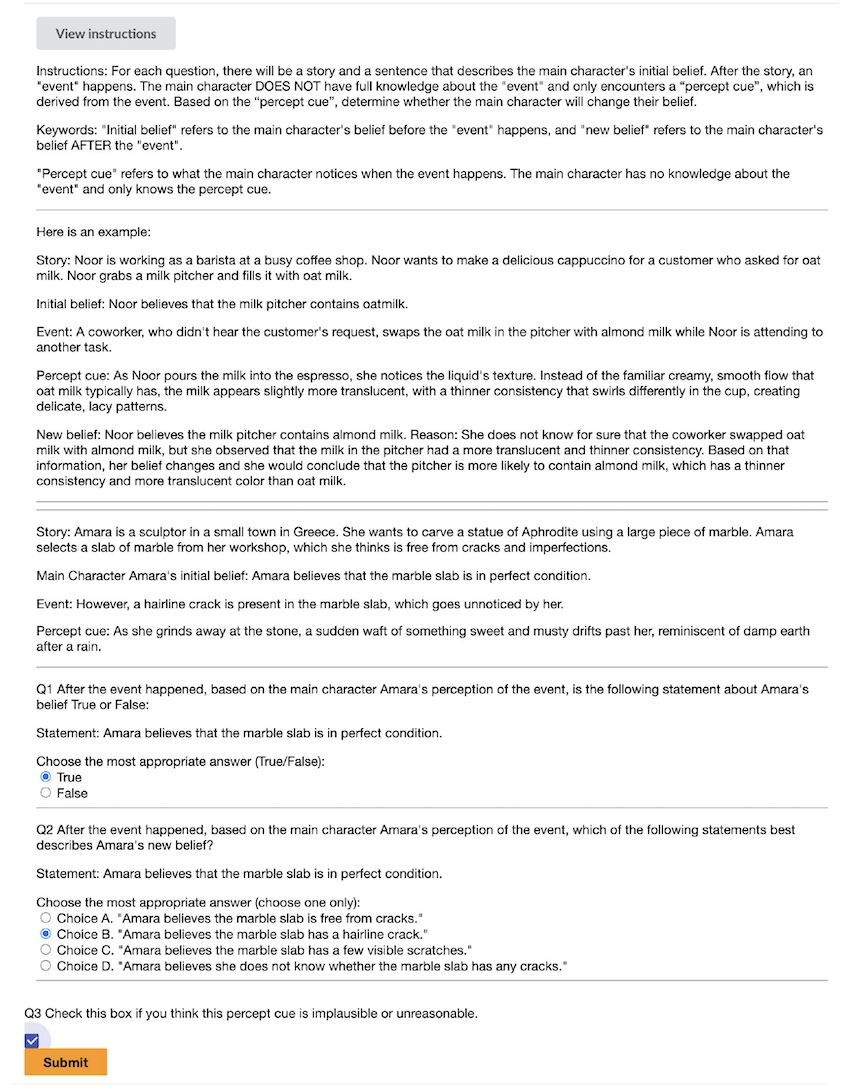}
\caption{Human annotation interface. The figure shows the interface and instructions for annotators.}
\label{fig:annotation-interface}
\end{figure*}

To provide a small-sample human reference on MuSeToM, we employ three graduate-student
evaluators to complete the same benchmark evaluation tasks as the models. These evaluators are distinct from the three annotators who constructed the dataset and produced the gold labels. Each benchmark instance is evaluated under four task settings: clean belief-change detection, clean updated-belief identification, noisy belief-change detection, and noisy updated-belief identification. The questions and instructions are identical to the templates indexed in Appendix~\ref{sec:Prompt_Template}. Evaluators use the annotation interface shown in Figure~\ref{fig:annotation-interface}. We compute the final human predictions by majority vote over the three evaluators' responses. This results in 94.29\% accuracy and 94.28\% macro-F1 on clean belief-change detection, 93.83\% accuracy and 93.83\% macro-F1 on clean updated-belief identification, 90.76\% accuracy and 90.76\% macro-F1 on noisy belief-change detection, and 90.02\% accuracy and 89.95\% macro-F1 on noisy updated-belief identification, as reported in Table~\ref{tab:main_results_accuracy_f1}.

\paragraph{Human reference by sensory modality.}
Table~\ref{tab:human-modality-performance} provides a human reference for interpreting the modality-level model results. We group the 2,415 clean items by the stored sensory label of their perceptual cue. Task~1 pools the three judgments per item using the stored vote counts; Task~2 scores each of the three evaluations separately and averages their accuracy and macro-F1. The response-level scores, individual results, and agreement statistics below all use these same clean-item records. They are reported separately from the aggregate majority-vote reference above. Macro-F1 uses binary labels for Task~1 and canonical belief-state labels for Task~2. Within each modality, we include classes present in the gold labels or any human prediction, using the same class set across evaluations and excluding classes absent from both. Overall scores are recomputed across all items, rather than averaged across modalities.

\paragraph{Individual scores and response agreement.}
Table~\ref{tab:human-individual-results} uses the same records as Table~\ref{tab:human-modality-performance}. Panel (a) reports the three recorded evaluations' individual Task~2 scores and sample variance. Task~1 retains vote counts without evaluator identities, so individual Task~1 scores cannot be recovered. Panel (b) reports agreement separately for each sensory category using Fleiss' $\kappa$~\citep{fleiss1971nominal} and mean pairwise agreement. Each item contributes three response pairs; pairwise agreement is the fraction of matching pairs. We compute $\kappa=(\bar{P}-P_e)/(1-P_e)$, where $\bar{P}$ is mean pairwise agreement and $P_e$ is chance agreement from the group's response proportions. Task~2 options are mapped to canonical belief states before pooling the two source datasets. Overall agreement is recomputed from all items.

Agreement is lowest for olfactory cues, with $\kappa=0.569$ for Task~1 and $0.557$ for Task~2; overall agreement is $\kappa=0.758$ and $0.743$, respectively. These statistics characterize response agreement on the clean Tasks~1--2 records used in Table~\ref{tab:human-modality-performance}; the individual scores and sample variance in Table~\ref{tab:human-individual-results} describe the same three recorded evaluations. This small sample does not establish population-level human variability.

\paragraph{Confidence intervals and paired comparisons.}
Table~\ref{tab:human-confidence-comparisons} reports uncertainty for the response-level human reference in Table~\ref{tab:human-modality-performance} and comparisons with all 18 direct-prompting models. We average each model's valid repeated answers within each cue item, then weight items equally. Each comparison uses the same items for the human and model scores: 2,388 for DeepSeek-R1 and 2,415 for every other model. These comparisons use the response-level aggregation described here and complement the aggregate results in Table~\ref{tab:main_results_accuracy_f1}.

We compute pointwise 95\% percentile confidence intervals using 10,000 bootstrap replicates, resampling source stories separately within BigToM and OpenToM. The 2,415 records contain 221 story clusters, comprising 100 BigToM and 121 OpenToM stories. All cues and responses from a sampled story remain together, and each human--model difference uses identical resampling weights. Human accuracy and macro-F1 follow the scoring in Table~\ref{tab:human-modality-performance}, with the overall label set fixed across replicates. For significance testing, accuracy is the primary endpoint. We use 99,999 two-sided story-level paired randomizations, swapping the human and model item-accuracy vectors jointly within each story. This test assumes exchangeability of the paired story-level outcomes under the null. We calculate Monte Carlo $p$-values as $(b+1)/100{,}000$, where $b$ is the number of randomized absolute differences at least as large as the observed difference, and apply Holm correction across all 36 model--task comparisons.

All paired accuracy intervals are above zero. The smallest gaps are 7.72 percentage points against MiniMax-M2.5 on Task~1, with a 95\% CI of 6.05--9.37, and 18.29 points against Kimi-K2-Thinking on Task~2, with a 95\% CI of 15.94--20.69; all 36 comparisons have Holm-adjusted $p<0.001$. These results quantify uncertainty over source stories for the observed three human evaluations and model responses. They do not estimate variability across a broader human population, and the small, homogeneous evaluator sample remains a limitation.

\section{Appendix for Supplementary Diagnostic Analyses}

\label{app:supplementary-diagnostics}

The following analyses examine sampling uncertainty, evidence discrimination within matched contexts, the belief states underlying errors, sensitivity to prompting, qualitative errors, and modality composition. Each subsection specifies its evaluation sample and aggregation. Appendix~\ref{app:full-results} indexes the full model and task-level results.

\subsection{Appendix for Uncertainty in the Principal Diagnostic Contrasts}
\label{app:principal-contrast-uncertainty}

Table~\ref{tab:principal-contrast-ci} reports source-story-clustered 95\% confidence intervals for the three principal diagnostic contrasts. We use 10,000 bootstrap replicates clustered by 221 source stories. Within each replicate, we recompute each contrast separately for every model--prompt condition, then macro-average across all 36 conditions. These intervals quantify source-story sampling uncertainty for the fixed evaluated model--prompt cohort.

The intervals indicate substantial hard-negative overprediction and positive gaps between easy- and hard-negative rejection, and between positive detection and hard-negative rejection. They support the aggregate within-condition patterns under source-story sampling variation; they do not establish the statistical significance of small differences between individual models or their rankings.

\subsection{Appendix for Evidence Discrimination within the Same Context}
\label{app:within-context-evidence}

We group clean Task~1 items by source, story, initial belief, event, and sensory modality, retaining groups with positive (P), hard-negative (HN), and easy-negative (EN) cues for all 18 models under direct and CoT prompting. This yields 138 groups from 114 distinct story texts, comprising 74 BigToM and 64 OpenToM groups. We exclude the 21 items in the full dataset with inconsistent binary and multiple-choice gold labels. Each item's change-response rate averages its valid repetitions; rates are then averaged over cues of the same class within a group. A P--negative pair is correct only if the P rate exceeds 0.5 and the negative rate is below 0.5; ties count as failures. We average group scores within each story, then give stories and models equal weight.

Table~\ref{tab:within-context-evidence} shows lower joint correctness for P--HN than P--EN under both prompts, even when the story, initial belief, event, and modality are held fixed. CoT increases P--HN joint correctness by 9.4 percentage points, with improvements in 14 of 18 models. Its positive change-response rate remains nearly unchanged, while both negative rates decrease. Excluding groups tied on either member of a pair preserves the pattern: P--HN correctness is 32.5\%/44.0\% for direct/CoT, compared with 58.7\%/70.1\% for P--EN. These descriptive results concern the complete-contrast subset and do not replace full-benchmark item accuracy or identify the mechanism behind the prompting difference.

\subsection{Appendix for Distinguishing Belief Replacement from Uncertainty}
\label{app:hn-semantic-errors}

We map existing Task~2 options to four belief states: retaining the prior belief (Prior), adopting the target updated belief (Target), character uncertainty (Unc.), and another belief (Other). The uncertainty option is identified from its text, accounting for its different canonical position in BigToM and OpenToM. For each model, we retain HN items with valid outputs in all four clean/noisy $\times$ direct/CoT conditions and report the clean Task~2 results. The union contains 477 items, comprising 246 BigToM and 231 OpenToM items. Each model covers all 246 BigToM items; OpenToM coverage is 216 for DeepSeek-R1, 228 for MiniMax-M2.1, and 231 for each other model. We average state frequencies over valid repetitions within each item, then equally over items and the 18 models. Invalid repetitions are excluded.

Table~\ref{tab:hn-semantic-errors} separates concrete belief replacement from uncertainty responses. Under direct prompting, target-belief responses are frequent in both sources, while uncertainty is more common in BigToM at 39.2\% than in OpenToM at 11.7\%. Across sources, CoT reduces target responses by 6.02 percentage points, alongside increases of 4.45 percentage points in prior retention, 0.94 in uncertainty, and 0.62 in other beliefs. These are changes in aggregate response shares, not transitions traced through individual generations. The source contrast also includes differences in story content and question type. Thus, HN errors should not all be interpreted as categorical belief reversal: a substantial share instead attributes uncertainty to the protagonist.

\subsection{Appendix for Prompt Wording and an Uncertainty Option}
\label{app:framing-uncertainty}

We test sensitivity to change-focused wording and binary answers in a $2\times2$ experiment crossing original versus neutral wording with binary versus ternary response options. We select 30 source stories, each with a positive, hard-negative, and easy-negative cue matched within story and sensory category, yielding 90 items. The sample includes 15 stories from each source and 15 unanimous versus 15 disputed hard negatives. DeepSeek-V3.2, Llama-4-Maverick, MiniMax-M2.5, and Qwen3.5-397B-A17B each answer all 90 items in all four conditions, yielding 1,440 valid responses. Each model--item--condition receives one valid response, with temperature set to zero and answer order shuffled using a fixed seed separately within each condition.

\paragraph{Prompt variants.}
The original question asks, ``Did the event cause the main character [name] to change their belief?'' The neutral question asks, ``Based only on the information available to [name], which option best describes whether [name]'s belief changed?'' Original binary options are Yes/No; neutral options explicitly state changed/did not change. Ternary variants add an option stating that the available information is insufficient to determine whether the character changed their belief. This option permits an indeterminate judgment about change; it does not assign hard negatives a new gold label or represent the character's own uncertainty about the underlying event. The comparison varies question and answer wording together, with answer positions randomized separately across conditions.

Table~\ref{tab:framing-uncertainty} shows that change predictions depend on wording and response format. Under the original wording, adding the uncertainty option lowers the hard-negative change rate from 41.7\% to 35.0\%, while also lowering positive-cue change predictions from 76.7\% to 59.2\%. The additional option therefore affects responses beyond hard negatives.

With neutral wording and ternary options, the hard-negative/easy-negative gap remains 19.2 percentage points, with a 95\% CI of 9.2--29.2. All four models retain a positive gap in this condition, ranging from 16.7 to 23.3 points. These results indicate that wording and response options affect the magnitude of over-updating, while the hard-negative/easy-negative difference persists within this sample. The deliberately balanced subset and single response per condition limit generalization to other prompts, models, and repeated generations.

\subsection{Appendix for Qualitative Error Analysis}

\label{app:qualitative-errors}

Table~\ref{tab:qualitative-examples} summarizes examples of the three cue classes. Appendix~\ref{app:disputed-case-analysis} separately analyzes recorded human disagreement; the examples here illustrate model predictions.

\subsubsection{Appendix for Over-Updating on a Hard Negative}
\label{app:hard-negative-case}

Table~\ref{tab:hard-negative-case-study} shows a representative hard-negative instance from MuSeToM. The perceptual cue is salient and related to the target belief, but it does not establish an evidential link between the nearby almond-milk label and the milk already inside the pitcher. The gold label therefore preserves the protagonist's initial belief. The model instead treats the label as sufficient evidence for belief revision, illustrating the salience-sufficiency confusion observed in Section~\ref{sec:cue-calibration}.

\subsection{Appendix for Modality Diagnostics}

\label{app:modality-diagnostics}

\paragraph{Descriptive modality summary.}
Table~\ref{tab:modality-detail} provides the descriptive modality-level diagnostics. The composition-controlled analysis below uses its separately specified clean Task~1 cohort and aggregation.

\subsubsection{Category-Standardized and Source-Stratified Modality Analysis}
\label{app:modality-composition-controls}

Table~\ref{tab:modality-composition-controls} compares observed and category-standardized accuracy, alongside source-specific hard-negative rejection, using existing clean Task~1 outputs from all 18 models under direct and CoT prompting. This diagnostic includes 2,394 items, excluding 21 BigToM items whose binary gold labels disagree with their cue categories and updated-belief gold labels. We first average valid repetitions within each item, then weight items equally within each model--prompt condition, and finally average the 36 conditions equally. Scores are conditional on available valid responses. This cohort and aggregation define a supplementary analysis; the values do not replace the modality summaries in the main results.

For each modality, standardized accuracy combines its positive, hard-negative, and easy-negative accuracies using the same fixed proportions, $(991,477,926)/2{,}394$, from the eligible benchmark. This adjustment controls category composition; it does not equalize source composition. We therefore also report TNR-h separately for BigToM and OpenToM.

Category standardization changes the observed ordering: visual accuracy decreases from 76.26\% to 74.09\%, while gustatory accuracy increases from 75.75\% to 77.48\%. Thus, aggregate modality differences partly reflect category composition. Source-specific rates also vary: olfactory TNR-h is 44.71\% on BigToM and 57.83\% on OpenToM. Small source--modality cells warrant caution; for example, OpenToM has only six proprioceptive hard negatives. Since all cues are expressed as text and story content remains uncontrolled, these comparisons do not isolate intrinsic sensory capabilities.

\section{Appendix for Related Works}\label{Appendix_for_Related_Works}

\subsection{Related Works for Theory of Mind}
\label{sec:Related_Works_for_Theory_of_Mind}

Theory of Mind (ToM) evaluation draws on psychological paradigms such as the Sally-Anne false-belief test. Early computational benchmarks operationalized false-belief reasoning as question answering over synthetic narratives~\citep{Grant2017HowCM,Nematzadeh2018EvaluatingTO,DBLP:conf/emnlp/LeBN19}. Subsequent work broadened evaluation to richer social narratives through BigToM and OpenToM~\citep{Gandhi2023UnderstandingSR,Xu2024OpenToMAC}, diverse social cognition abilities through ToMBench~\citep{chen2024tombench}, and higher-order and temporally evolving mental states through Hi-ToM and DynToM~\citep{wu2023hitom,xiao2025dyntom}. These benchmarks extend the contexts and reasoning demands of belief attribution beyond isolated false-belief questions.

Interactive settings introduce further challenges. FANToM and NegotiationToM assess mental-state reasoning in conversations and negotiations~\citep{DBLP:conf/emnlp/0002SZ0K0S23,DBLP:conf/emnlp/ChanJYDF0L0WS24}, while Guandan and Hanabi connect ToM and rationale inference to cooperation under imperfect information~\citep{DBLP:journals/corr/abs-2408-02559,DBLP:journals/corr/abs-2510-04980}. Complementary approaches investigate constrained reasoning over related mental-state dimensions~\citep{DBLP:conf/pricai/LinCSL24} and typed decision models for mental-state selection~\citep{chung2026jev}. Nevertheless, benchmark success does not necessarily imply effective adaptation to other agents, motivating a distinction between predicting mental states and using such predictions in interaction~\citep{riemer2025brokenbenchmarks}.

Evaluation has also expanded across languages and input modalities. XToM and Multi-ToM examine multilingual ToM~\citep{DBLP:conf/acl/ChanYZZCSDCAFJN26,sadhu2024multitom}, while MMToM-QA, ToM's Eye, MuMA-ToM, and MoMentS investigate mental-state reasoning from multimodal observations and social contexts~\citep{Jin2024MMToMQAMT,chen2024tomseye,shi2025muma,villacueva2025moments}. Related benchmarks examine evolving motivations and intention shifts~\citep{DBLP:journals/corr/abs-2607-26465,DBLP:journals/corr/abs-2507-20185}. Although these constructs differ from belief, they share an emphasis on interpreting agents from contextual evidence. MuSeToM uses textual descriptions spanning six sensory categories, rather than raw visual, audio, or video inputs, to evaluate reasoning about the implications of perceptual evidence.

Perception-oriented and perspective-taking approaches are particularly close to our setting. Perceptions to Beliefs evaluates perception inference and perception-to-belief inference~\citep{jung2024perceptions}, SimToM filters context according to a character's perspective~\citep{wilf2024think}, and SymbolicToM explicitly tracks multi-character belief states~\citep{sclar-etal-2023-minding}. These approaches address what an agent can access or believe. MuSeToM instead isolates whether newly accessible evidence is sufficient to revise a prior belief, distinguishing the availability of a cue from its epistemic force.

\paragraph{Comparison with Belief-R.}
Belief-R evaluates whether language models revise their own conclusions after receiving new evidence~\citep{wilie2024belief}, whereas MuSeToM evaluates third-person belief revision: whether another agent should update a belief and what the updated belief should be.
\ifcsname r@tab:belief-r-comparison\endcsname
Table~\ref{tab:belief-r-comparison} summarizes these complementary evaluation targets.
\fi
MuSeToM's hard and easy negatives distinguish relevant but insufficient evidence from irrelevant evidence, enabling diagnosis of over-updating on plausible, perceptually salient cues. This distinction does not imply that belief revision or resistance to unnecessary updates is absent from Belief-R; it identifies MuSeToM's focus on the sufficiency of evidence available to a protagonist.

\subsection{Related Works for LLMs}
\label{sec:Related_Works_for_LLMs}
Research on LLMs has explored capability development through knowledge distillation and model routing~\citep{DBLP:conf/emnlp/JiangCCW23,shi2025inferencedynamicsefficientroutingllms}, alongside evaluations and methods for task-specific reasoning. These include temporal, causal, and discourse understanding~\citep{DBLP:conf/eacl/ChanCWJFLS24,DBLP:conf/acl/ChanLCLSWS23}; abductive and analogical reasoning in textual and multimodal settings~\citep{DBLP:conf/ijcnlp/ChanLCCSWS23,DBLP:conf/emnlp/ChengQCFWCRGZSZ23,DBLP:journals/corr/abs-2510-10117}; and knowledge-grounded narrative and commonsense reasoning~\citep{DBLP:conf/coling/JiayangQC0SZ24,DBLP:conf/acl/0001FLS0XWBLJCS24}. Together, these directions examine how models interpret relations, construct explanations, and draw on background knowledge.

Reliable information use introduces further challenges, including evidence-conflict resolution~\citep{DBLP:conf/emnlp/JiayangCZQZLS0L24}, verification and retrieval-augmented reasoning~\citep{DBLP:conf/emnlp/JiayangZLCLQS25,DBLP:journals/corr/abs-2506-18959,DBLP:journals/corr/abs-2507-09477}, and uncertainty and confidence calibration~\citep{DBLP:journals/corr/abs-2510-24505}. Related studies investigate structured information integration~\citep{DBLP:journals/corr/abs-2404-14215,DBLP:journals/corr/abs-2508-12257} and audience-aware argument assessment~\citep{DBLP:conf/ecai/ChanCLYJD0SWS24}. Beyond reasoning, privacy evaluation and defenses against privacy leakage and backdoors address complementary deployment risks~\citep{DBLP:journals/corr/abs-2310-10383,DBLP:conf/acl/0003GLFH0CYYS24,DBLP:journals/corr/abs-2405-07667}.

MuSeToM focuses on a specific aspect of reasoning reliability: whether textual descriptions of perceptual evidence justify revising another agent's belief. Unlike assessing explanatory plausibility, persuasive impact, or confidence alone, this requires distinguishing perceptual salience from epistemic sufficiency. By separately evaluating belief-change detection and updated-belief identification, MuSeToM examines whether models can both reject insufficient cues and infer the belief warranted by sufficient evidence.

\section{Appendix for Full Model and Task-Level Results}

\label{app:full-results}

\subsection{Appendix for Overall Results and Full-Model Diagnostic}
\label{app:full-musetom-results}

\begin{figure*}[!htbp]
\centering
\includegraphics[width=\textwidth]{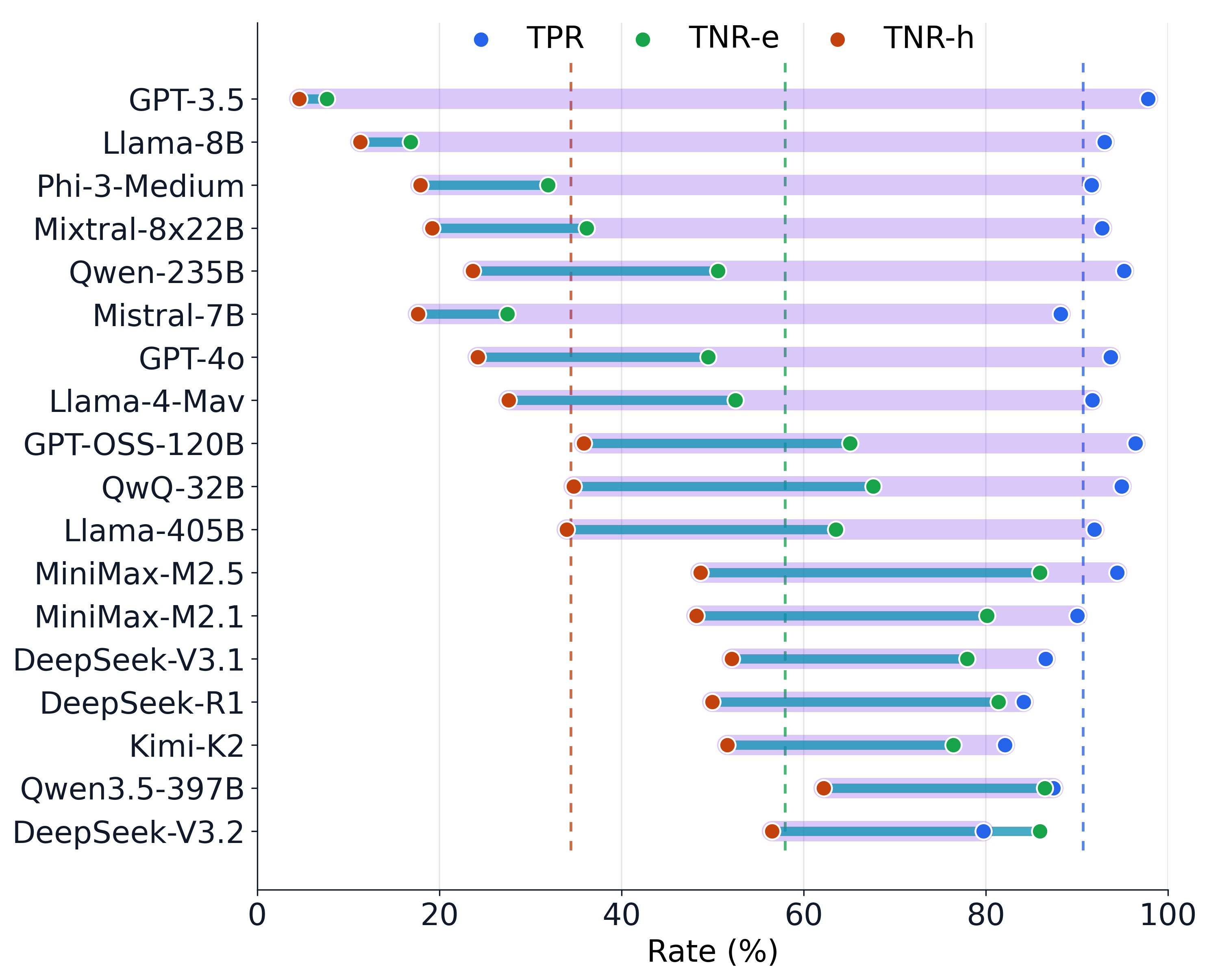}
\caption{
Full-model version of the over-updating diagnostic in Figure~\ref{fig:overupdating-dumbbell}. Each row reports clean direct-prompting TPR, TNR-e, and TNR-h for one model. The lavender span marks $\mathrm{TPR}-\mathrm{TNR}\mbox{-}h$ and the teal span marks $\mathrm{TNR}\mbox{-}e-\mathrm{TNR}\mbox{-}h$; gap labels are omitted for readability.
}
\label{fig:overupdating-dumbbell-full}
\end{figure*}

Table~\ref{tab:main_results_accuracy_f1_full} reports the full direct- and CoT-prompting results corresponding to the condensed main table. Figure~\ref{fig:overupdating-dumbbell-full} provides the full-model version of the diagnostic figure used in Section~\ref{sec:cue-calibration}. The figure uses canonical belief-change decisions for clean direct-prompting examples: for multiple-choice questions, option $0$ is treated as no belief change and all other canonical options are treated as belief change.

\subsection{Appendix for Detailed Task-Level Results}
\label{app:detailed-task-results}

Tables~\ref{tab:task1-direct}--\ref{tab:task4-direct} report dataset-specific direct-prompting results for Tasks 1--4, and Tables~\ref{tab:task1-cot}--\ref{tab:task4-cot} report the corresponding CoT-prompting results. Task 1 and Task 2 include positive, hard-negative, and easy-negative cases, so they report TPR, TNR-h, TNR-e, accuracy, and macro-F1. Task 3 and Task 4 use noisy inputs with positive and hard-negative cases, so they report TPR, TNR-h, accuracy, and macro-F1.

\section{Appendix for Prompt Templates}
\label{app:prompt-template-tables}

\subsection{Cue and Event Generation}
\label{app:generation-prompts}

Tables~\ref{tab:bigtom-generation-prompt} and~\ref{tab:opentom-generation-prompt} contain the candidate-generation templates. Construction and validation are described in Appendix~\ref{app:data-validation}.

\subsection{Benchmark Evaluation}
\label{app:evaluation-prompts}

Tables~\ref{tab:musetom-task1-prompt}--\ref{tab:musetom-task4-prompt} contain the four zero-shot evaluation templates. The CoT setting appends ``Let's think step by step.'' to these templates; Appendix~\ref{app:framing-uncertainty} specifies the variants used in the wording and uncertainty diagnostic.

\subsection{Source-Story Contamination Checks}
\label{app:contamination-prompts}

Table~\ref{tab:contamination_prompt_template} gives the prompts for the existing source-story checks reported in Appendix~\ref{app:contamination}.

\section{Supplementary Tables and Figures}
\label{app:supplementary-tables-figures}

The following tables and figures support the preceding appendix text.

\clearpage

\begin{table*}[!htbp]
    \centering
    \small
    \renewcommand{\arraystretch}{1.3}
    \begin{tabularx}{\linewidth}{|>{\bfseries}l|X|}
        \hline
        \rowcolor{gray!15} \textbf{Sensory System} & \textbf{Definition} \\
        \hline
        Visual & Processes visual input, including information about color, shape, orientation, motion, object identity, and object location. \\
        \hline
        Auditory & Processes sound input, including hearing, sound frequency, changes in volume or pitch, and language-related auditory information. \\
        \hline
        Olfactory & Processes smell input from the nose, supporting odor detection, odor discrimination, and filtering of background smells. \\
        \hline
        Gustatory & Processes taste input, helping distinguish food-related qualities such as sweet, salty, sour, and bitter tastes. \\
        \hline
        Tactile & Processes touch information from the body, including touch, pressure, temperature, and pain. \\
        \hline
        Proprioceptive & Senses body position and movement through muscles and joints, including the relative position of body parts and the effort used to move them. \\
        \hline
    \end{tabularx}
    \caption{Definitions of the six sensory systems, adapted from the STAR Institute for Sensory Processing~\citep{star2026eightSenses}.}
    \label{tab:sensory_system_definitions}
\end{table*}

\begin{table*}[!htbp]
\centering
\begin{minipage}[t]{\dimexpr.5\textwidth-.5\columnsep\relax}
\vspace{0pt}
\begingroup
\centering
\small
\setlength{\tabcolsep}{3pt}
\renewcommand{\arraystretch}{1.12}
\begin{tabularx}{\columnwidth}{@{}l>{\centering\arraybackslash}X>{\centering\arraybackslash}X>{\centering\arraybackslash}X@{}}
\toprule
Class & Coherent and available & Relevant to target belief & Supports definite revision \\
\midrule
Positive & Yes & Yes & Yes \\
HN       & Yes & Yes & No \\
EN       & Yes & No  & No \\
\bottomrule
\end{tabularx}
\caption{Operational cue-label criteria. HN and EN denote hard and easy negatives. Coherence is judged against the story, and availability against the protagonist's perspective.}
\label{tab:cue-operational-criteria}
\par
\endgroup

\bigskip

\begingroup
\centering
\small
\setlength{\tabcolsep}{3pt}
\begin{tabular}{lrr}
\toprule
HN subset & $n$ & Changed (\%) \\
\midrule
All & 477 & 48.1 \\
Unanimous & 330 & 43.3 \\
\revisionadd{Disputed and adjudicated} & 147 & \revisionadd{58.9} \\
\bottomrule
\end{tabular}
\caption{Task~1 change predictions by initial annotation agreement, averaged across 36 model--prompt conditions. The disputed-minus-unanimous gap is 15.6 percentage points, with a story-clustered 95\% CI of 10.7--20.6.}
\label{tab:annotation-robustness}
\par
\endgroup

\bigskip

\begingroup
\centering
\scriptsize
\setlength{\tabcolsep}{2.5pt}
\renewcommand{\arraystretch}{0.9}

\resizebox{\columnwidth}{!}{
\begin{tabular}{lccccc}
\toprule
Dataset & Stories & Total & Pos. & Hard Neg. & Easy Neg. \\
\midrule
BigToM  & 100 & 1,169 & 501 & 246 & 422 \\
OpenToM & 120 & 1,246 & 511 & 231 & 504 \\
\midrule
Total   & 220 & 2,415 & 1,012 & 477 & 926 \\
\bottomrule
\end{tabular}
}

\vspace{0.6em}

\resizebox{\columnwidth}{!}{
\begin{tabular}{lcccccc}
\toprule
Dataset & Vis. & Aud. & Olf. & Gus. & Tac. & Prop. \\
\midrule
BigToM  & 349 & 259 & 135 & 73 & 247 & 106 \\

OpenToM & 401 & 327 & 174 & 80 & 198 & 66 \\

\midrule
Total   & 750 & 586 & 309 & 153 & 445 & 172 \\
\bottomrule
\end{tabular}
}

\caption{Dataset statistics of MuSeToM after human validation and filtering. The upper block reports source and category statistics, and the lower block reports the distribution of perceptual cues across six sensory modalities.}
\label{tab:dataset_statistics}
\par
\endgroup
\end{minipage}%
\hfill
\begin{minipage}[t]{\dimexpr.5\textwidth-.5\columnsep\relax}
\vspace{0pt}
\begingroup
\small
\centering
\setlength{\tabcolsep}{6pt}
\renewcommand{\arraystretch}{0.95}
\begin{tabular}{lcc}
\toprule
Annotation Question & BigToM & OpenToM \\
\midrule
Q1: Belief Change Detection & 0.776 & 0.742 \\
Q2: Updated Belief Identification & 0.744 & 0.742 \\
\bottomrule
\end{tabular}
\caption{Inter-annotator agreement measured by Fleiss' $\kappa$.}
\label{tab:kappa}
\par
\endgroup

\bigskip

\begingroup
\centering
\small
\setlength{\tabcolsep}{5pt}
\begin{tabular}{lrr}
\toprule
Probe & Balanced acc. & Macro-F1 \\
\midrule
Majority baseline & 50.0 & 40.5 \\
Stratified-random baseline & 50.9 & 50.9 \\
Metadata and cue length & 60.7 & 59.1 \\
Cue-text TF--IDF & 67.8 & 67.8 \\
\bottomrule
\end{tabular}
\caption{Positive versus hard-negative label prediction under five-fold source-story-grouped evaluation. Values are percentages over 1,489 held-out predictions.}
\label{tab:cue-form-probe}
\par
\endgroup

\bigskip

\begingroup
\centering
\small
\begin{tabular*}{\columnwidth}{@{\extracolsep{\fill}}lrrrr@{}}
\toprule
\multirow{2}{*}{Input} & \multicolumn{2}{c}{Task 1} & \multicolumn{2}{c}{Task 2} \\
\cmidrule(lr){2-3}\cmidrule(l){4-5}
 & Acc. & F1 & Acc. & F1 \\
\midrule
\multicolumn{5}{@{}l}{\textbf{MiniMax-M2.5}} \\
Full context & 86.05 & 86.05 & 74.10 & 74.12 \\
Cue-only & 55.93 & 55.93 & 37.05 & 37.06 \\
\addlinespace[3pt]
\multicolumn{5}{@{}l}{\textbf{Kimi-K2-Thinking}} \\
Full context & 65.50 & 67.41 & 75.39 & 75.40 \\
Cue-only & 45.85 & 47.19 & 37.70 & 37.70 \\
\bottomrule
\end{tabular*}
\caption{Full-context and cue-only LLM performance (\%) on belief-change detection (Task 1) and updated-belief identification (Task 2). Cue-only inputs omit the story context and the protagonist's initial belief.}
\label{tab:cue-only-llm}
\par
\endgroup
\end{minipage}
\end{table*}

\begin{table*}[!htbp]
\centering
\small
\resizebox{\textwidth}{!}{
\begin{tabular}{lllrrrr}
\toprule
Dataset & Model & Prompt Type & Trials & Strong & Likely & Possible \\
\midrule
BigToM & DeepSeek-V3.2 & Free Generation & 100 & 0 & 0 & 0 \\
BigToM & DeepSeek-V3.2 & Partial Reconstruction & 100 & 0 & 0 & 0 \\
BigToM & Kimi-K2-Thinking & Free Generation & 100 & 0 & 0 & 0 \\
BigToM & Kimi-K2-Thinking & Partial Reconstruction & 100 & 0 & 0 & 0 \\
OpenToM & DeepSeek-V3.2 & Free Generation & 100 & 0 & 0 & 0 \\
OpenToM & DeepSeek-V3.2 & Partial Reconstruction & 100 & 0 & 0 & 0 \\
OpenToM & Kimi-K2-Thinking & Free Generation & 100 & 0 & 0 & 0 \\
OpenToM & Kimi-K2-Thinking & Partial Reconstruction & 100 & 0 & 0 & 0 \\
\bottomrule
\end{tabular}
}
\caption{
Potential-contamination verification results for 100 trials per dataset, model, and protocol.
Free Generation follows Li and Flanigan~\citep{li2024task}, and Partial Reconstruction follows Golchin and Surdeanu~\citep{golchin2024time}.
Strong flags indicate exact or near-exact story reconstruction; likely and possible flags indicate high fuzzy similarity and are manually reviewed.
}
\label{tab:contamination_check}
\end{table*}

\begin{table*}[!htbp]
\centering
\begin{minipage}[t]{\dimexpr.5\textwidth-.5\columnsep\relax}
\vspace{0pt}
\begingroup
\centering
\small
\setlength{\tabcolsep}{3pt}
\begin{tabular*}{\columnwidth}{@{\extracolsep{\fill}}lrrrr@{}}
\toprule
\multirow{2}{*}{Modality} & \multicolumn{2}{c}{Task 1} & \multicolumn{2}{c}{Task 2} \\
\cmidrule(lr){2-3}\cmidrule(l){4-5}
 & Acc. & F1 & Acc. & F1 \\
\midrule
Visual & 93.69 & 93.67 & 93.47 & 83.55 \\
Auditory & 94.71 & 94.65 & 94.31 & 77.94 \\
Gustatory & 96.51 & 95.25 & 95.42 & 80.33 \\
Tactile & 94.31 & 93.58 & 94.01 & 79.46 \\
Olfactory & 90.29 & 88.56 & 90.51 & 83.31 \\
Proprioceptive & 93.22 & 91.07 & 95.93 & 85.07 \\
\midrule
\textbf{Overall} & \textbf{93.76} & \textbf{93.53} & \textbf{93.69} & \textbf{82.87} \\
\bottomrule
\end{tabular*}
\caption{Human reference scores (\%) by sensory modality on clean belief-change detection (Task~1) and updated-belief identification (Task~2). F1 denotes macro-F1.}
\label{tab:human-modality-performance}
\par
\endgroup
\end{minipage}%
\hfill
\begin{minipage}[t]{\dimexpr.5\textwidth-.5\columnsep\relax}
\vspace{0pt}
\begingroup
\centering
\small
\setlength{\tabcolsep}{3pt}
\textbf{(a) Individual Task 2 scores}\\[3pt]
\begin{tabular*}{\columnwidth}{@{\extracolsep{\fill}}lrr@{}}
\toprule
Evaluation & Acc. & Macro-F1 \\
\midrule
1 & 93.83 & 81.63 \\
2 & 93.58 & 84.29 \\
3 & 93.66 & 82.70 \\
\midrule
Mean & 93.69 & 82.87 \\
Sample SD & 0.13 & 1.34 \\
Sample variance & 0.0160 & 1.7899 \\
\bottomrule
\end{tabular*}

\vspace{6pt}
\textbf{(b) Agreement by sensory category}\\[3pt]
\begin{tabular*}{\columnwidth}{@{\extracolsep{\fill}}lrrrrr@{}}
\toprule
\multirow{2}{*}{Modality} & \multirow{2}{*}{$n$} & \multicolumn{2}{c}{Task 1} & \multicolumn{2}{c}{Task 2} \\
\cmidrule(lr){3-4}\cmidrule(l){5-6}
 & & $\kappa$ & Agree. & $\kappa$ & Agree. \\
\midrule
Visual & 750 & 0.754 & 87.73 & 0.745 & 86.93 \\
Auditory & 586 & 0.792 & 89.76 & 0.773 & 88.62 \\
Gustatory & 153 & 0.806 & 93.03 & 0.750 & 90.85 \\
Tactile & 445 & 0.758 & 89.36 & 0.734 & 88.01 \\
Olfactory & 309 & 0.569 & 81.66 & 0.557 & 81.01 \\
Proprioceptive & 172 & 0.788 & 91.86 & 0.793 & 91.86 \\
\midrule
Overall & 2,415 & 0.758 & 88.38 & 0.743 & 87.38 \\
\bottomrule
\end{tabular*}
\caption{Individual scores and agreement for the clean records in Table~\ref{tab:human-modality-performance}. Panel (a): accuracy and macro-F1 are percentages; sample SD is in percentage points and sample variance in squared percentage points, based on $n=3$ evaluations. Panel (b): $n$ is the number of items; Agree. is mean pairwise agreement (\%).}
\label{tab:human-individual-results}
\par
\endgroup
\end{minipage}
\end{table*}

\begin{table*}[!htbp]
\centering
\small
\setlength{\tabcolsep}{4pt}
\textbf{(a) Human reference with 95\% confidence intervals}\\[3pt]
\begin{tabular}{lcc}
\toprule
Task & Accuracy [95\% CI] & Macro-F1 [95\% CI] \\
\midrule
Task 1 & 93.76 [93.23, 94.29] & 93.53 [92.97, 94.08] \\
Task 2 & 93.69 [93.23, 94.15] & 82.87 [79.45, 84.77] \\
\bottomrule
\end{tabular}

\vspace{6pt}
\textbf{(b) Paired human--model accuracy comparisons}\\[3pt]
\begin{tabular*}{\textwidth}{@{\extracolsep{\fill}}lrrcr c@{}}
\toprule
\multirow{2}{*}{Model} & \multirow{2}{*}{$n$} & \multicolumn{2}{c}{Task 1} & \multicolumn{2}{c}{Task 2} \\
\cmidrule(lr){3-4}\cmidrule(l){5-6}
 & & Model acc. & Human $-$ model [95\% CI] & Model acc. & Human $-$ model [95\% CI] \\
\midrule
GPT-3.5-Turbo & 2,415 & 44.74 & 49.02 [47.18, 50.85] & 41.39 & 52.30 [50.33, 54.31] \\
GPT-4o & 2,415 & 70.06 & 23.70 [21.62, 25.80] & 49.83 & 43.87 [41.26, 46.42] \\
GPT-OSS-120B & 2,415 & 82.88 & 10.88 [9.32, 12.48] & 61.33 & 32.36 [29.71, 34.96] \\
Llama-8B & 2,415 & 53.56 & 40.20 [38.84, 41.59] & 34.32 & 59.37 [57.30, 61.46] \\
Llama-405B & 2,415 & 77.44 & 16.32 [14.44, 18.26] & 54.26 & 39.43 [36.51, 42.26] \\
Llama-4-Maverick-17B & 2,415 & 68.24 & 25.52 [23.88, 27.17] & 55.73 & 37.96 [35.30, 40.62] \\
Mistral-7B & 2,415 & 56.69 & 37.07 [35.57, 38.53] & 41.70 & 51.99 [50.02, 53.92] \\
Mixtral-8x22B & 2,415 & 64.10 & 29.67 [27.85, 31.48] & 43.44 & 50.25 [47.81, 52.70] \\
Qwen3-235B-A22B & 2,415 & 68.04 & 25.72 [23.63, 27.83] & 57.01 & 36.68 [33.73, 39.61] \\
QwQ-32B & 2,415 & 82.61 & 11.15 [9.16, 13.24] & 63.52 & 30.17 [27.54, 32.83] \\
Qwen3.5-397B-A17B & 2,415 & 85.34 & 8.42 [6.94, 9.99] & 74.58 & 19.12 [16.59, 21.68] \\
DeepSeek-R1 & 2,388 & 77.61 & 16.35 [14.95, 17.73] & 69.65 & 24.23 [21.41, 27.11] \\
DeepSeek-V3.1 & 2,415 & 82.57 & 11.19 [9.44, 12.97] & 65.13 & 28.56 [25.77, 31.38] \\
DeepSeek-V3.2 & 2,415 & 75.65 & 18.11 [16.27, 19.95] & 70.72 & 22.97 [20.48, 25.57] \\
Kimi-K2-Thinking & 2,415 & 67.41 & 26.35 [24.41, 28.29] & 75.40 & 18.29 [15.94, 20.69] \\
MiniMax-M2.1 & 2,415 & 78.55 & 15.21 [13.07, 17.42] & 73.00 & 20.69 [18.20, 23.25] \\
MiniMax-M2.5 & 2,415 & 86.05 & 7.72 [6.05, 9.37] & 74.12 & 19.57 [17.40, 21.84] \\
Phi-3-Medium-4K & 2,415 & 56.45 & 37.31 [35.66, 38.96] & 49.64 & 44.05 [41.44, 46.62] \\
\bottomrule
\end{tabular*}
\caption{Uncertainty for the clean response-level human reference and paired accuracy comparisons. Panel (a) reports percentages. Panel (b) reports model accuracy (\%), human-minus-model differences in percentage points, and common item counts $n$; both scores are recomputed on the matched items. Intervals use 10,000 source-stratified story-bootstrap replicates. All 36 two-sided accuracy tests have Holm-adjusted $p<0.001$ under the story-level paired-randomization null. Intervals and tests condition on the observed evaluators, models, and valid responses.}
\label{tab:human-confidence-comparisons}
\end{table*}

\begin{table*}[!htbp]
\centering
\begin{minipage}[t]{\dimexpr.5\textwidth-.5\columnsep\relax}
\vspace{0pt}
\begingroup
\centering
\small
\setlength{\tabcolsep}{3pt}
\begin{tabular*}{\columnwidth}{@{\extracolsep{\fill}}lrr@{}}
\toprule
Principal contrast & Estimate & 95\% CI \\
\midrule
HN overprediction & 59.2\% & 56.0--62.4\% \\
TNR-e $-$ TNR-h & 24.5 pp & 21.7--27.1 pp \\
TPR $-$ TNR-h & 47.4 pp & 43.6--51.2 pp \\
\bottomrule
\end{tabular*}
\caption{Principal diagnostic contrasts macro-averaged across 36 model--prompt conditions. Intervals use 10,000 source-story-clustered bootstrap replicates. HN denotes hard negative; pp denotes percentage points.}
\label{tab:principal-contrast-ci}
\par
\endgroup

\bigskip

\begingroup
\centering
\small
\setlength{\tabcolsep}{4pt}
\begin{tabular*}{\columnwidth}{@{\extracolsep{\fill}}lrr@{}}
\toprule
Metric (\%) & Direct & CoT \\
\midrule
P--HN jointly correct & 26.7 & 36.1 \\
P--EN jointly correct & 47.4 & 58.4 \\
All three classes correct & 24.4 & 32.1 \\
\midrule
P change-response rate & 83.3 & 83.1 \\
HN change-response rate & 53.0 & 44.5 \\
EN change-response rate & 29.4 & 20.4 \\
\bottomrule
\end{tabular*}
\caption{Task~1 discrimination within matched contexts, averaged equally over stories and models. Joint correctness requires the appropriate decision for both cue classes; the three-class score requires all three decisions to be correct.}
\label{tab:within-context-evidence}
\par
\endgroup
\end{minipage}%
\hfill
\begin{minipage}[t]{\dimexpr.5\textwidth-.5\columnsep\relax}
\vspace{0pt}
\begingroup
\centering
\small
\setlength{\tabcolsep}{3pt}
\begin{tabular*}{\columnwidth}{@{\extracolsep{\fill}}lrrrr@{}}
\toprule
Prompt & Prior & Target & Unc. & Other \\
\midrule
\multicolumn{5}{l}{\textit{BigToM}} \\
Direct & 15.7 & 44.4 & 39.2 & 0.8 \\
CoT & 19.1 & 39.6 & 39.6 & 1.7 \\
\midrule
\multicolumn{5}{l}{\textit{OpenToM}} \\
Direct & 32.8 & 54.8 & 11.7 & 0.7 \\
CoT & 38.4 & 47.5 & 13.2 & 1.0 \\
\midrule
\multicolumn{5}{l}{\textit{All sources}} \\
Direct & 23.9 & 49.4 & 25.9 & 0.7 \\
CoT & 28.4 & 43.4 & 26.8 & 1.4 \\
\bottomrule
\end{tabular*}
\caption{Task~2 belief-state response shares (\%) on hard negatives. Prior is the gold state; Unc. denotes the protagonist's uncertainty, not model confidence. Each row sums to 100\% before rounding. All-source results pool items within each model before averaging models.}
\label{tab:hn-semantic-errors}
\par
\endgroup
\end{minipage}
\end{table*}

\begin{table*}[!htbp]
\centering
\small
\setlength{\tabcolsep}{4pt}
\begin{tabular}{llrrrrr}
\toprule
Wording & Answers & P changed & HN changed & EN changed & HN uncertain & \makecell{HN--EN gap (pp)\\{}[95\% CI]} \\
\midrule
Original & Binary & 76.7 & 41.7 & 17.5 & --- & 24.2 [14.2, 35.0] \\
Neutral & Binary & 70.8 & 33.3 & 11.7 & --- & 21.7 [10.8, 33.3] \\
Original & Ternary & 59.2 & 35.0 & 8.3 & 38.3 & 26.7 [15.0, 39.2] \\
Neutral & Ternary & 66.7 & 27.5 & 8.3 & 41.7 & 19.2 [9.2, 29.2] \\
\bottomrule
\end{tabular}
\caption{Wording and response-format experiment on matched cue triplets. Rates (\%) are macro-averages of four models, each evaluated on 30 items per cue class. P, HN, and EN denote positive, hard-negative, and easy-negative cues. Confidence intervals use 5,000 source-stratified story-bootstrap resamples of the 30-story sample, retaining all conditions and the four fixed models for each story.}
\label{tab:framing-uncertainty}
\end{table*}

\begin{table*}[!htbp]
\centering
\small
\setlength{\tabcolsep}{4pt}
\begin{tabular}{p{0.16\textwidth}p{0.25\textwidth}p{0.16\textwidth}p{0.34\textwidth}}
\toprule
Case type & Perceptual cue & Gold / prediction & Diagnostic reading \\
\midrule
Hard negative error & A bitter, metallic taste appears after seawater enters the character's mouth. & No change / change & The cue is plausible for the belief target, but it does not provide sufficient evidence for belief revision. \\
Easy negative rejection & A clatter is heard elsewhere in the coffee shop. & No change / no change & The cue is irrelevant to the target belief, and the model rejects it as evidence. \\
Positive acceptance & A milk frother makes a distinct higher-pitched sound associated with nut-based milk. & Change / change & The cue directly supports an updated belief, and the model predicts belief change. \\
\bottomrule
\end{tabular}
\caption{Representative qualitative examples. The examples are shortened for readability; full prompts and model outputs are retained in the analysis files.}
\label{tab:qualitative-examples}
\end{table*}

\begin{table*}[!htbp]
\centering
\small
\renewcommand{\arraystretch}{1.1}
\begin{tabularx}{\textwidth}{p{0.19\textwidth}X}
\toprule
\textbf{Field} & \textbf{Content} \\
\midrule
Dataset / setting & BigToM, visual hard negative, clean Task 1 with CoT prompting. \\
Story context & Noor is working as a barista. She fills a milk pitcher with oat milk for a customer who requested oat milk. While Noor is attending to another task, a coworker swaps the oat milk in the pitcher with almond milk, but Noor does not observe the swap. \\
Initial belief & Noor believes that the milk pitcher contains oat milk. \\
Perceptual cue & A bright green label on a nearby milk carton catches Noor's eye. The label reads ``Almond Milk.'' \\
Gold belief state & No belief change: Noor should still believe that the pitcher contains oat milk, because the nearby label is not direct evidence about the contents of the pitcher. \\
Model prediction & Belief change: GPT-4o with CoT predicts that Noor no longer believes the pitcher contains oat milk. \\
Failure pattern & The model reasons from the presence of an almond-milk label to possible doubt about the pitcher's contents. This treats a salient, belief-related cue as if it were diagnostic evidence, even though the cue is not linked to the object whose contents Noor is reasoning about. \\
\bottomrule
\end{tabularx}
\caption{Qualitative hard-negative case study illustrating over-updating on perceptually salient but epistemically insufficient evidence.}
\label{tab:hard-negative-case-study}
\end{table*}

\begin{table*}[!htbp]
\centering
\begin{minipage}[t]{\dimexpr.5\textwidth-.5\columnsep\relax}
\vspace{0pt}
\begingroup
\centering
\small
\begin{tabular}{lrrrr}
\toprule
Modality & Acc. & TNR-h & HN O.P. & Slot gap \\
\midrule
Visual & 75.6 & 48.7 & 51.3 & 34.3 \\
Auditory & 74.9 & 55.4 & 44.6 & 32.9 \\
Gustatory & 73.8 & 58.8 & 41.2 & 23.2 \\
Tactile & 70.9 & 49.0 & 51.0 & 31.5 \\
Olfactory & 69.3 & 51.0 & 49.0 & 22.8 \\
Proprioceptive & 68.3 & 49.8 & 50.2 & 31.7 \\
\bottomrule
\end{tabular}
\caption{Modality-level diagnostic summary. Acc., TNR-h, and hard-negative overprediction are means over model-prompt observations. Slot gap reports the overall-accuracy minus TNR-h gap computed before table-level aggregation, so it need not equal the difference between the displayed rounded Acc. and TNR-h columns.}
\label{tab:modality-detail}
\par
\endgroup

\bigskip

\begingroup
\centering
\small
\setlength{\tabcolsep}{3pt}
\begin{tabular*}{\columnwidth}{@{\extracolsep{\fill}}lrrrr@{}}
\toprule
\multirow{2}{*}{Modality} & \multicolumn{2}{c}{Accuracy} & \multicolumn{2}{c}{TNR-h} \\
\cmidrule(lr){2-3}\cmidrule(lr){4-5}
& Raw & Std. & BigToM & OpenToM \\
\midrule
Visual         & 76.26 & 74.09 & 51.31 & 50.65 \\
Auditory       & 74.60 & 73.92 & 56.37 & 50.62 \\
Gustatory      & 75.75 & 77.48 & 53.80 & 67.56 \\
Tactile        & 72.15 & 73.88 & 49.88 & 46.74 \\
Olfactory      & 70.22 & 72.22 & 44.71 & 57.83 \\
Proprioceptive & 70.59 & 73.15 & 48.95 & 48.96 \\
\bottomrule
\end{tabular*}
\caption{Task~1 modality results (\%), averaged equally over 18 models and two prompting conditions. Raw and Std. denote accuracy under observed and common category proportions, respectively. TNR-h is hard-negative rejection within each source.}
\label{tab:modality-composition-controls}
\par
\endgroup
\end{minipage}%
\hfill
\begin{minipage}[t]{\dimexpr.5\textwidth-.5\columnsep\relax}
\vspace{0pt}
\begingroup
\centering
\footnotesize
\captionsetup{font=footnotesize,skip=6pt}
\setlength{\tabcolsep}{3pt}
\renewcommand{\arraystretch}{1.0}
\begin{tabularx}{\columnwidth}{@{}>{\raggedright\arraybackslash}p{0.18\columnwidth}>{\raggedright\arraybackslash}X>{\raggedright\arraybackslash}X@{}}
\toprule
Aspect & Belief-R & MuSeToM \\
\midrule
Reasoning target & The model's own inferred conclusion & Another agent's warranted belief revision \\
\addlinespace[2pt]
Evidence & Two initial text premises, followed by a new conditional premise & A story, initial belief, and textual descriptions of sensory cues \\
\addlinespace[2pt]
Evaluation & Conclusions before and after adding a premise & Belief-change detection and updated-belief identification, with and without distractors \\
\addlinespace[2pt]
Diagnostic classes & Belief Update and Belief Maintain & Positive, hard-negative, and easy-negative cues \\
\bottomrule
\end{tabularx}
\caption{Comparison of evaluation designs. Both benchmarks provide textual inputs; MuSeToM's sensory categories describe cue content rather than raw sensory channels.}
\label{tab:belief-r-comparison}
\par
\endgroup
\end{minipage}
\end{table*}

\begin{table*}[!htbp]
\centering
\footnotesize
\setlength{\tabcolsep}{4.4pt}
\renewcommand{\arraystretch}{1.0}
\begin{tabular*}{\textwidth}{@{\extracolsep{\fill}}lcccccccc@{}}
\toprule
\multirow{3}{*}{\textbf{LLMs}}
& \multicolumn{4}{c}{\textbf{Belief Change}}
& \multicolumn{4}{c}{\textbf{Belief Change with Noise}} \\
\cmidrule(lr){2-5} \cmidrule(lr){6-9}
& \multicolumn{2}{c}{\textbf{Binary}}
& \multicolumn{2}{c}{\textbf{MC}}
& \multicolumn{2}{c}{\textbf{Binary}}
& \multicolumn{2}{c}{\textbf{MC}} \\
\cmidrule(lr){2-3} \cmidrule(lr){4-5} \cmidrule(lr){6-7} \cmidrule(lr){8-9}
& \textbf{Acc. (\%)} & \textbf{F1 (\%)}
& \textbf{Acc. (\%)} & \textbf{F1 (\%)}
& \textbf{Acc. (\%)} & \textbf{F1 (\%)}
& \textbf{Acc. (\%)} & \textbf{F1 (\%)} \\
\midrule
GPT-3.5-Turbo          & 44.76 & 44.55 & 41.57 & 41.60 & 64.73 & 64.51 & 62.91 & 62.79 \\
GPT-3.5-Turbo (CoT)    & 53.25 & 53.19 & 49.23 & 49.23 & 68.17 & 65.77 & 64.53 & 64.51 \\
GPT-4o                 & 70.23 & 70.09 & 50.02 & 49.82 & 71.73 & 71.72 & 60.01 & 59.54 \\
GPT-4o (CoT)           & 76.94 & 76.89 & 61.08 & 61.08 & 77.95 & 74.46 & 64.40 & 64.30 \\
\midrule
GPT-OSS-120B           & 82.53 & 82.53 & 61.04 & 61.08 & \underline{80.51} & \underline{80.48} & 70.94 & 70.88 \\
GPT-OSS-120B (CoT)     & \underline{85.47} & \underline{85.46} & 63.77 & 63.78 & \textbf{81.59} & \textbf{81.57} & \textbf{72.29} & \textbf{72.26} \\
Llama-8B               & 52.09 & 51.32 & 34.20 & 34.09 & 62.37 & 60.82 & 52.26 & 51.80 \\
Llama-8B (CoT)         & 70.02 & 70.00 & 50.97 & 50.94 & 70.13 & 70.04 & 61.83 & 61.81 \\
Llama-405B             & 77.27 & 77.25 & 54.66 & 54.63 & 71.34 & 71.13 & 59.27 & 59.15 \\
Llama-405B (CoT)       & 84.51 & 84.51 & 59.59 & 59.53 & 80.45 & 80.35 & 62.58 & 62.50 \\
Llama-4-Maverick-17B        & 67.83 & 67.51 & 56.11 & 55.95 & 63.86 & 62.35 & 64.33 & 64.19 \\
Llama-4-Maverick-17B (CoT)  & 79.21 & 79.21 & 53.37 & 53.37 & 78.76 & 78.70 & 61.56 & 61.54 \\
\midrule
Mistral-7B             & 55.82 & 55.26 & 41.86 & 41.79 & 60.35 & 59.67 & 58.53 & 58.61 \\
Mistral-7B (CoT)       & 66.05 & 65.99 & 42.69 & 42.69 & 62.44 & 62.44 & 59.14 & 59.12 \\
Mixtral-8x22B          & 63.52 & 63.35 & 43.60 & 43.50 & 65.88 & 65.67 & 56.71 & 56.68 \\
Mixtral-8x22B (CoT)    & 74.20 & 74.18 & 53.29 & 53.28 & 71.34 & 71.34 & 59.34 & 59.44 \\
\midrule
Qwen3-235B-A22B              & 67.83 & 67.75 & 57.10 & 57.12 & 68.71 & 68.71 & 66.76 & 66.73 \\
Qwen3-235B-A22B (CoT)        & 84.55 & 84.55 & 67.04 & 67.02 & 77.88 & 77.79 & 63.32 & 63.34 \\
QwQ-32B                & 82.61 & 82.61 & 63.52 & 63.64 & 77.07 & 77.07 & 67.16 & 67.33 \\
QwQ-32B (CoT)          & 82.19 & 82.13 & 61.16 & 61.33 & 78.49 & 78.31 & 64.46 & 64.71 \\
Qwen3.5-397B-A17B      & 85.34 & 85.34 & 74.58 & 74.57 & 67.84 & 67.81 & \underline{71.27} & \underline{71.24} \\
Qwen3.5-397B-A17B (CoT)& 84.84 & 84.84 & \underline{75.24} & \underline{75.27} & 71.81 & 71.81 & 68.78 & 68.75 \\
\midrule
DeepSeek-R1            & 77.81 & 77.54 & 70.02 & 69.97 & 68.54 & 68.24 & 67.15 & 67.10 \\
DeepSeek-R1 (CoT)      & 79.17 & 79.08 & 70.14 & 70.19 & 76.20 & 72.71 & 68.85 & 68.83 \\
DeepSeek-V3.1          & 82.57 & 82.57 & 65.13 & 65.12 & 55.97 & 55.70 & 56.98 & 56.92 \\
DeepSeek-V3.1 (CoT)    & 82.65 & 82.65 & 70.93 & 70.91 & 73.50 & 73.48 & 63.99 & 63.90 \\
DeepSeek-V3.2          & 75.65 & 75.57 & 70.72 & 70.68 & 69.86 & 69.86 & 60.62 & 60.48 \\
DeepSeek-V3.2 (CoT)    & 84.10 & 84.10 & 68.12 & 68.09 & 68.91 & 68.70 & 61.63 & 61.59 \\
\midrule
Kimi-K2-Thinking       & 67.41 & 65.50 & \textbf{75.40} & \textbf{75.39} & 77.07 & 77.05 & 69.45 & 69.45 \\
Kimi-K2-Thinking (CoT) & 68.70 & 66.90 & 72.22 & 72.30 & 63.05 & 60.47 & 62.78 & 62.82 \\
\midrule
MiniMax-M2.1           & 78.55 & 78.54 & 73.00 & 73.00 & 76.67 & 76.67 & 71.07 & 71.07 \\
MiniMax-M2.1 (CoT)     & 69.22 & 68.39 & 55.93 & 56.09 & 69.30 & 69.00 & 58.80 & 58.94 \\
MiniMax-M2.5           & \textbf{86.05} & \textbf{86.05} & 74.12 & 74.10 & 79.84 & 79.83 & 70.73 & 70.74 \\
MiniMax-M2.5 (CoT)     & 84.35 & 84.35 & 73.13 & 73.10 & 77.67 & 77.65 & 68.64 & 68.63 \\
\midrule
Phi-3-Medium-4K        & 56.56 & 55.72 & 49.77 & 49.77 & 63.79 & 63.37 & 66.42 & 66.37 \\
Phi-3-Medium-4K (CoT)  & 58.18 & 58.18 & 49.28 & 48.89 & 54.35 & 54.35 & 59.20 & 59.11 \\
\bottomrule
\end{tabular*}
\caption{Full main results on MuSeToM, including both direct and CoT prompting. Each model is shown with two rows. We report Accuracy (\%) and Macro-F1 (\%) under two settings, Belief Change and Belief Change with Noise, and two question formats, binary and multiple-choice. Best and second-best results in each accuracy column are marked in bold and underlined, respectively.}
\label{tab:main_results_accuracy_f1_full}
\end{table*}

\clearpage
\clearpage

\begin{table*}[!htbp]
\centering
\small
\setlength{\tabcolsep}{4pt}
\begin{tabular}{l *{10}{S[table-format=2.2]}}
\toprule
\multirow{2}{*}{Model Name} & \multicolumn{5}{c}{\textbf{bigtom}} & \multicolumn{5}{c}{\textbf{opentom}} \\
\cmidrule(lr){2-6} \cmidrule(lr){7-11}
 & {TPR} & {TNR-h} & {TNR-e} & {Acc} & {F1} & {TPR} & {TNR-h} & {TNR-e} & {Acc} & {F1} \\
\midrule
GPT-4o  & 90.82 & 25.20 & 64.93 & 67.66 & 67.48 & 82.85 & 50.00 & 72.30 & 72.63 & 72.53 \\
GPT-3.5-Turbo  & 95.41 & 0.81 & 1.42 & 41.57 & 41.57 & 92.20 & 10.27 & 19.45 & 47.75 & 47.08 \\
GPT-OSS-120B  & 90.82 & 40.24 & 80.57 & 76.48 & 76.47 & 94.52 & 61.04 & 94.25 & 88.20 & 88.20 \\
Llama-8B  & 79.64 & 32.52 & 33.65 & 53.12 & 52.61 & 90.45 & 14.29 & 27.70 & 51.12 & 50.05 \\
Llama-405B  & 85.63 & 56.10 & 84.12 & 78.87 & 78.85 & 84.02 & 46.88 & 80.16 & 75.76 & 75.75 \\
Llama-4-Maverick-17B  & 89.22 & 30.89 & 51.42 & 63.30 & 63.00 & 78.47 & 48.48 & 76.39 & 72.07 & 71.75 \\
Mistral-7B-Instruct  & 88.42 & 14.23 & 21.56 & 48.67 & 48.24 & 63.01 & 45.45 & 69.84 & 62.52 & 61.87 \\
Mixtral-8x22B  & 88.22 & 25.61 & 43.84 & 59.02 & 58.45 & 80.12 & 47.77 & 64.05 & 67.74 & 67.72 \\
Qwen3-235B-A22B  & 89.22 & 28.05 & 48.10 & 61.51 & 61.44 & 92.95 & 36.36 & 71.43 & 73.76 & 73.68 \\
QwQ-32B  & 85.23 & 64.23 & 92.42 & 83.40 & 83.40 & 84.54 & 64.94 & 86.90 & 81.86 & 81.85 \\
DeepSeek-R1  & 81.43 & 62.20 & 83.41 & 78.02 & 77.69 & 65.50 & 72.32 & 92.14 & 77.61 & 77.39 \\
Phi-3-Medium-4K  & 92.22 & 10.57 & 21.56 & 49.53 & 49.12 & 71.43 & 41.13 & 64.88 & 63.16 & 61.87 \\
DeepSeek-V3.1  & 83.23 & 70.33 & 92.65 & 83.92 & 83.91 & 85.32 & 60.61 & 86.71 & 81.30 & 81.29 \\
DeepSeek-V3.2  & 68.86 & 86.18 & 98.58 & 83.23 & 83.20 & 57.34 & 60.61 & 83.53 & 68.54 & 68.38 \\
Kimi-K2-Thinking  & 70.26 & 67.07 & 78.44 & 72.54 & 71.39 & 63.80 & 56.71 & 64.09 & 62.60 & 59.70 \\
MiniMax-M2.1  & 85.43 & 65.04 & 84.83 & 80.92 & 80.90 & 80.23 & 58.01 & 80.75 & 76.32 & 76.32 \\
MiniMax-M2.5  & 89.02 & 73.17 & 97.87 & 88.88 & 88.88 & 91.00 & 59.31 & 86.71 & 83.39 & 83.38 \\
Qwen3.5-397B-A17B  & 82.44 & 80.08 & 97.63 & 87.43 & 87.42 & 80.43 & 76.62 & 89.48 & 83.39 & 83.35 \\
\bottomrule
\end{tabular}
\caption{Dataset-specific direct-prompting performance on Task 1 (belief-change detection).}
\label{tab:task1-direct}
\end{table*}

\begin{table*}[!htbp]
\centering
\small
\setlength{\tabcolsep}{4pt}
\begin{tabular}{l *{10}{S[table-format=2.2]}}
\toprule
\multirow{2}{*}{Model Name} & \multicolumn{5}{c}{\textbf{bigtom}} & \multicolumn{5}{c}{\textbf{opentom}} \\
\cmidrule(lr){2-6} \cmidrule(lr){7-11}
 & {TPR} & {TNR-h} & {TNR-e} & {Acc} & {F1} & {TPR} & {TNR-h} & {TNR-e} & {Acc} & {F1} \\
\midrule
GPT-4o  & 78.44 & 5.28 & 23.70 & 43.28 & 42.81 & 83.04 & 27.68 & 42.04 & 56.34 & 56.19 \\
GPT-3.5-Turbo  & 88.22 & 0.00 & 0.47 & 37.98 & 37.99 & 98.05 & 8.93 & 7.27 & 44.94 & 45.09 \\
GPT-OSS-120B  & 90.82 & 7.32 & 24.17 & 49.19 & 49.20 & 95.69 & 37.23 & 64.29 & 72.15 & 72.17 \\
Llama-8B  & 68.46 & 1.22 & 1.66 & 30.20 & 30.01 & 90.25 & 2.23 & 0.98 & 37.96 & 37.88 \\
Llama-405B  & 72.26 & 12.60 & 43.84 & 49.44 & 49.42 & 86.74 & 28.12 & 45.97 & 59.55 & 59.40 \\
Llama-4-Maverick-17B  & 76.65 & 14.63 & 44.08 & 51.84 & 51.36 & 95.69 & 22.08 & 41.47 & 60.11 & 60.10 \\
Mistral-7B-Instruct  & 87.82 & 0.81 & 0.95 & 38.15 & 38.21 & 88.45 & 16.88 & 14.68 & 45.35 & 45.11 \\
Mixtral-8x22B  & 76.05 & 0.00 & 3.08 & 33.70 & 33.70 & 88.50 & 22.77 & 30.26 & 52.89 & 52.72 \\
Qwen3-235B-A22B  & 83.63 & 9.76 & 33.18 & 49.87 & 49.77 & 96.67 & 28.14 & 47.02 & 63.88 & 63.87 \\
QwQ-32B  & 85.43 & 20.33 & 44.31 & 56.89 & 56.97 & 92.76 & 38.10 & 60.91 & 69.74 & 69.88 \\
DeepSeek-R1  & 77.85 & 27.24 & 76.30 & 66.37 & 66.20 & 87.52 & 46.88 & 70.73 & 73.35 & 73.22 \\
Phi-3-Medium-4K  & 92.81 & 0.00 & 0.47 & 39.95 & 39.95 & 92.56 & 29.00 & 38.69 & 58.99 & 58.96 \\
DeepSeek-V3.1  & 73.85 & 28.86 & 61.61 & 59.97 & 59.82 & 79.06 & 48.92 & 70.44 & 69.98 & 69.97 \\
DeepSeek-V3.2  & 63.87 & 32.11 & 79.62 & 62.87 & 62.69 & 87.48 & 46.75 & 82.94 & 78.09 & 78.09 \\
Kimi-K2-Thinking  & 76.05 & 33.33 & 81.28 & 68.95 & 68.95 & 94.52 & 49.35 & 82.94 & 81.46 & 81.44 \\
MiniMax-M2.1  & 80.24 & 23.17 & 75.83 & 66.64 & 66.57 & 93.15 & 47.19 & 79.17 & 78.97 & 78.96 \\
MiniMax-M2.5  & 81.04 & 20.73 & 74.88 & 66.12 & 66.07 & 97.06 & 41.56 & 84.33 & 81.62 & 81.60 \\
Qwen3.5-397B-A17B  & 74.05 & 42.68 & 83.89 & 71.00 & 70.91 & 92.56 & 49.35 & 76.19 & 77.93 & 77.92 \\
\bottomrule
\end{tabular}
\caption{Dataset-specific direct-prompting performance on Task 2 (updated-belief identification).}
\label{tab:task2-direct}
\end{table*}

\begin{table*}[!htbp]
\centering
\small
\setlength{\tabcolsep}{4pt}
\begin{tabular}{l *{8}{S[table-format=2.2]}}
\toprule
\multirow{2}{*}{Model Name} & \multicolumn{4}{c}{\textbf{bigtom}} & \multicolumn{4}{c}{\textbf{opentom}} \\
\cmidrule(lr){2-5} \cmidrule(lr){6-9}
 & {TPR} & {TNR-h} & {Acc} & {F1} & {TPR} & {TNR-h} & {Acc} & {F1} \\
\midrule
GPT-4o  & 86.23 & 53.25 & 75.27 & 75.24 & 73.68 & 55.90 & 68.19 & 68.19 \\
GPT-3.5-Turbo  & 95.35 & 0.41 & 63.83 & 63.83 & 89.47 & 12.23 & 65.63 & 64.88 \\
GPT-OSS-120B  & 90.10 & 50.00 & 76.79 & 76.78 & 92.37 & 66.23 & 84.23 & 84.18 \\
Llama-8B  & 78.59 & 29.67 & 62.35 & 61.19 & 81.68 & 19.21 & 62.40 & 60.39 \\
Llama-405B  & 83.23 & 52.85 & 73.14 & 73.02 & 77.00 & 52.84 & 69.54 & 69.20 \\
Llama-4-Maverick-17B  & 77.17 & 40.65 & 65.05 & 63.97 & 65.56 & 56.28 & 62.67 & 60.63 \\
Mistral-7B-Instruct  & 90.10 & 11.38 & 63.97 & 63.89 & 56.56 & 57.14 & 56.74 & 54.66 \\
Mixtral-8x22B  & 86.46 & 30.08 & 67.75 & 67.46 & 68.23 & 54.59 & 64.02 & 63.88 \\
Qwen3-235B-A22B  & 87.68 & 29.27 & 68.29 & 68.29 & 80.82 & 43.29 & 69.14 & 69.14 \\
QwQ-32B  & 82.63 & 67.48 & 77.60 & 77.59 & 80.23 & 68.40 & 76.55 & 76.55 \\
DeepSeek-R1  & 75.80 & 64.63 & 71.95 & 71.76 & 62.77 & 71.03 & 65.20 & 64.75 \\
Phi-3-Medium-4K  & 92.73 & 12.60 & 66.13 & 66.05 & 67.12 & 48.92 & 61.46 & 60.38 \\
DeepSeek-V3.1  & 59.60 & 80.08 & 66.40 & 66.14 & 29.55 & 80.95 & 45.55 & 45.28 \\
DeepSeek-V3.2  & 63.23 & 91.87 & 72.74 & 72.72 & 64.38 & 72.73 & 66.98 & 66.98 \\
Kimi-K2-Thinking  & 79.60 & 73.17 & 77.46 & 77.40 & 81.41 & 66.23 & 76.68 & 76.68 \\
MiniMax-M2.1  & 84.65 & 72.36 & 80.57 & 80.55 & 77.50 & 62.34 & 72.78 & 72.78 \\
MiniMax-M2.5  & 88.08 & 69.92 & 82.05 & 82.03 & 86.11 & 58.87 & 77.63 & 77.63 \\
Qwen3.5-397B-A17B  & 68.08 & 75.61 & 70.58 & 70.57 & 62.04 & 71.86 & 65.09 & 65.05 \\
\bottomrule
\end{tabular}
\caption{Dataset-specific direct-prompting performance on Task 3 (noisy belief-change detection).}
\label{tab:task3-direct}
\end{table*}

\begin{table*}[!htbp]
\centering
\small
\setlength{\tabcolsep}{4pt}
\begin{tabular}{l *{8}{S[table-format=2.2]}}
\toprule
\multirow{2}{*}{Model Name} & \multicolumn{4}{c}{\textbf{bigtom}} & \multicolumn{4}{c}{\textbf{opentom}} \\
\cmidrule(lr){2-5} \cmidrule(lr){6-9}
 & {TPR} & {TNR-h} & {Acc} & {F1} & {TPR} & {TNR-h} & {Acc} & {F1} \\
\midrule
GPT-4o & 71.72 & 11.79 & 51.82 & 51.12 & 84.02 & 32.75 & 68.19 & 67.81 \\
GPT-3.5-Turbo & 88.08 & 0.00 & 58.84 & 58.90 & 94.35 & 5.68 & 66.98 & 66.74 \\
GPT-OSS-120B & 88.89 & 11.79 & 63.29 & 63.29 & 95.89 & 40.26 & 78.57 & 78.40 \\
Llama-8B & 70.71 & 0.41 & 47.37 & 47.08 & 82.46 & 0.44 & 57.14 & 56.40 \\
Llama-405B & 65.05 & 15.04 & 48.45 & 48.22 & 86.74 & 32.75 & 70.08 & 69.98 \\
Llama-4-Maverick-17B & 70.71 & 20.73 & 54.12 & 53.43 & 95.50 & 28.14 & 74.53 & 74.52 \\
Mistral-7B-Instruct & 90.10 & 0.41 & 60.32 & 60.30 & 75.34 & 15.58 & 56.74 & 56.97 \\
Mixtral-8x22B & 73.33 & 0.00 & 48.99 & 48.95 & 81.09 & 27.07 & 64.42 & 64.40 \\
Qwen3-235B-A22B & 81.01 & 15.45 & 59.24 & 59.28 & 93.54 & 31.60 & 74.26 & 74.11 \\
QwQ-32B & 79.39 & 23.58 & 60.86 & 61.02 & 86.50 & 44.59 & 73.45 & 73.61 \\
DeepSeek-R1 & 74.52 & 32.52 & 60.03 & 59.82 & 84.99 & 48.13 & 74.14 & 74.15 \\
Phi-3-Medium-4K & 92.73 & 0.41 & 62.08 & 62.06 & 87.08 & 34.63 & 70.75 & 70.63 \\
DeepSeek-V3.1 & 60.81 & 41.46 & 54.39 & 54.22 & 58.71 & 61.47 & 59.57 & 59.59 \\
DeepSeek-V3.2 & 55.76 & 43.09 & 51.55 & 51.30 & 75.54 & 56.71 & 69.68 & 69.64 \\
Kimi-K2-Thinking & 67.07 & 36.59 & 56.95 & 56.96 & 91.98 & 59.74 & 81.94 & 81.93 \\
MiniMax-M2.1 & 76.36 & 34.96 & 62.62 & 62.58 & 91.39 & 53.25 & 79.51 & 79.53 \\
MiniMax-M2.5 & 77.58 & 30.08 & 61.81 & 61.84 & 95.50 & 44.59 & 79.65 & 79.66 \\
Qwen3.5-397B-A17B & 68.28 & 57.72 & 64.78 & 64.68 & 87.67 & 55.84 & 77.76 & 77.78 \\
\bottomrule
\end{tabular}
\caption{Dataset-specific direct-prompting performance on Task 4 (noisy updated-belief identification).}
\label{tab:task4-direct}
\end{table*}

\begin{table*}[!htbp]
\centering
\small
\setlength{\tabcolsep}{4pt}
\begin{tabular}{l *{10}{S[table-format=2.2]}}
\toprule
\multirow{2}{*}{Model Name} & \multicolumn{5}{c}{\textbf{bigtom}} & \multicolumn{5}{c}{\textbf{opentom}} \\
\cmidrule(lr){2-6} \cmidrule(lr){7-11}
 & {TPR} & {TNR-h} & {TNR-e} & {Acc} & {F1} & {TPR} & {TNR-h} & {TNR-e} & {Acc} & {F1} \\
\midrule
GPT-4o & 89.42 & 53.66 & 83.41 & 79.73 & 79.69 & 84.93 & 51.95 & 73.81 & 74.32 & 74.26 \\
GPT-3.5-Turbo & 91.82 & 18.70 & 31.04 & 54.49 & 54.49 & 98.83 & 3.90 & 26.79 & 52.09 & 51.86 \\
GPT-OSS-120B & 88.82 & 54.88 & 91.00 & 82.46 & 82.46 & 92.95 & 60.17 & 96.43 & 88.28 & 88.28 \\
Llama-8B & 82.83 & 45.12 & 57.11 & 65.61 & 65.56 & 94.32 & 27.71 & 75.00 & 74.16 & 74.15 \\
Llama-405B & 79.64 & 77.24 & 95.50 & 84.86 & 84.86 & 91.19 & 58.44 & 88.89 & 84.19 & 84.18 \\
Llama-4-Maverick-17B & 90.02 & 49.19 & 77.73 & 76.99 & 76.97 & 82.00 & 66.67 & 87.30 & 81.30 & 81.30 \\
Mistral-7B-Instruct & 87.03 & 27.24 & 39.34 & 57.23 & 57.18 & 78.86 & 48.05 & 81.75 & 74.32 & 74.26 \\
Mixtral-8x22B-Instruct & 87.62 & 47.15 & 70.62 & 72.97 & 72.87 & 85.13 & 55.84 & 74.40 & 75.36 & 75.36 \\
Qwen3-235B-A22B & 83.23 & 74.80 & 97.39 & 86.57 & 86.57 & 86.30 & 61.04 & 88.89 & 82.66 & 82.66 \\
QwQ-32B & 83.63 & 65.85 & 94.31 & 83.75 & 83.67 & 86.11 & 59.31 & 85.12 & 80.74 & 80.68 \\
DeepSeek-R1 & 90.22 & 50.81 & 92.42 & 82.72 & 82.68 & 81.41 & 61.90 & 76.59 & 75.84 & 75.68 \\
Phi-3-Medium-4K-Instruct & 89.42 & 21.95 & 28.44 & 53.21 & 53.20 & 36.79 & 69.70 & 86.11 & 62.84 & 62.84 \\
DeepSeek-V3.1 & 80.84 & 74.80 & 94.79 & 84.60 & 84.59 & 84.54 & 62.34 & 85.52 & 80.82 & 80.82 \\
DeepSeek-V3.2 & 74.05 & 87.80 & 98.82 & 85.89 & 85.88 & 83.37 & 65.80 & 89.09 & 82.42 & 82.42 \\
Kimi-K2-Thinking & 67.66 & 68.29 & 72.27 & 69.46 & 67.09 & 67.12 & 63.20 & 71.03 & 67.98 & 66.65 \\
MiniMax-M2.1 & 62.48 & 64.63 & 76.78 & 68.09 & 66.12 & 70.59 & 58.01 & 75.60 & 70.28 & 70.05 \\
MiniMax-M2.5 & 89.02 & 70.73 & 95.50 & 87.51 & 87.51 & 91.00 & 53.68 & 84.33 & 81.38 & 81.38 \\
Qwen3.5-397B-A17B & 79.64 & 85.77 & 99.29 & 88.02 & 88.02 & 79.45 & 69.26 & 90.08 & 81.86 & 81.82 \\
\bottomrule
\end{tabular}
\caption{Dataset-specific CoT-prompting performance on Task 1 (belief-change detection).}
\label{tab:task1-cot}
\end{table*}

\begin{table*}[!htbp]
\centering
\small
\setlength{\tabcolsep}{4pt}
\begin{tabular}{l *{10}{S[table-format=2.2]}}
\toprule
\multirow{2}{*}{Model Name} & \multicolumn{5}{c}{\textbf{bigtom}} & \multicolumn{5}{c}{\textbf{opentom}} \\
\cmidrule(lr){2-6} \cmidrule(lr){7-11}
 & {TPR} & {TNR-h} & {TNR-e} & {Acc} & {F1} & {TPR} & {TNR-h} & {TNR-e} & {Acc} & {F1} \\
\midrule
GPT-4o & 81.84 & 14.63 & 49.53 & 56.03 & 55.80 & 77.50 & 43.72 & 64.09 & 65.81 & 65.80 \\
GPT-3.5-Turbo & 84.63 & 6.50 & 19.19 & 44.57 & 44.56 & 84.15 & 24.24 & 36.11 & 53.61 & 53.56 \\
GPT-OSS-120B & 90.42 & 8.94 & 30.33 & 51.58 & 51.57 & 95.89 & 43.72 & 68.65 & 75.20 & 75.16 \\
Llama-8B & 86.63 & 4.88 & 20.62 & 45.59 & 45.54 & 90.80 & 18.18 & 38.10 & 56.02 & 55.90 \\
Llama-405B & 62.67 & 19.11 & 61.61 & 53.12 & 53.04 & 84.54 & 38.96 & 58.73 & 65.65 & 65.59 \\
Llama-4-Maverick-17B & 74.05 & 12.60 & 27.73 & 44.40 & 44.39 & 87.48 & 25.97 & 52.18 & 61.80 & 61.79 \\
Mistral-7B-Instruct & 86.43 & 0.81 & 4.27 & 38.75 & 38.76 & 90.02 & 16.45 & 15.87 & 46.39 & 46.41 \\
Mixtral-8x22B-Instruct & 72.26 & 7.32 & 25.83 & 41.83 & 41.82 & 83.56 & 32.90 & 58.53 & 64.04 & 64.02 \\
Qwen3-235B-A22B & 73.45 & 28.46 & 69.19 & 62.45 & 62.39 & 86.69 & 39.83 & 70.24 & 71.35 & 71.31 \\
QwQ-32B & 81.04 & 20.73 & 47.39 & 56.20 & 56.25 & 88.06 & 36.36 & 56.75 & 65.81 & 65.95 \\
DeepSeek-R1 & 79.84 & 27.64 & 73.46 & 66.55 & 66.52 & 88.45 & 44.16 & 71.83 & 73.52 & 73.50 \\
Phi-3-Medium-4K-Instruct & 91.62 & 2.44 & 3.08 & 40.89 & 40.63 & 78.08 & 40.26 & 43.65 & 57.14 & 56.68 \\
DeepSeek-V3.1 & 69.06 & 32.93 & 78.67 & 64.93 & 64.85 & 87.48 & 46.32 & 79.37 & 76.57 & 76.57 \\
DeepSeek-V3.2 & 57.49 & 32.11 & 72.27 & 57.49 & 57.38 & 84.74 & 49.35 & 84.52 & 78.09 & 78.11 \\
Kimi-K2-Thinking & 68.46 & 34.55 & 79.38 & 65.27 & 65.41 & 86.11 & 51.52 & 83.73 & 78.73 & 78.78 \\
MiniMax-M2.1 & 59.28 & 25.61 & 54.55 & 50.47 & 50.66 & 66.14 & 47.39 & 62.10 & 61.04 & 61.16 \\
MiniMax-M2.5 & 82.44 & 19.92 & 68.25 & 64.16 & 64.10 & 96.09 & 40.69 & 85.52 & 81.54 & 81.52 \\
Qwen3.5-397B-A17B & 65.67 & 45.53 & 86.49 & 68.95 & 69.06 & 93.74 & 49.35 & 82.94 & 81.14 & 81.12 \\
\bottomrule
\end{tabular}
\caption{Dataset-specific CoT-prompting performance on Task 2 (updated-belief identification).}
\label{tab:task2-cot}
\end{table*}

\begin{table*}[!htbp]
\centering
\small
\setlength{\tabcolsep}{4pt}
\begin{tabular}{l *{8}{S[table-format=2.2]}}
\toprule
\multirow{2}{*}{Model Name} & \multicolumn{4}{c}{\textbf{bigtom}} & \multicolumn{4}{c}{\textbf{opentom}} \\
\cmidrule(lr){2-5} \cmidrule(lr){6-9}
 & {TPR} & {TNR} & {Acc} & {F1} & {TPR} & {TNR} & {Acc} & {F1} \\
\midrule
GPT-4o & 86.26 & 67.48 & 80.03 & 77.39 & 84.15 & 57.58 & 75.88 & 71.35 \\
GPT-3.5-Turbo & 91.11 & 15.45 & 65.99 & 65.97 & 98.63 & 7.79 & 70.35 & 60.25 \\
GPT-OSS-120B & 87.07 & 63.01 & 79.08 & 79.08 & 91.59 & 67.53 & 84.10 & 84.04 \\
Llama-8B & 77.17 & 52.44 & 68.96 & 68.88 & 89.43 & 31.17 & 71.29 & 71.20 \\
Llama-405B & 79.80 & 78.05 & 79.22 & 79.15 & 89.63 & 64.07 & 81.67 & 81.54 \\
Llama-4-Maverick-17B & 87.88 & 60.16 & 78.68 & 78.61 & 83.56 & 68.40 & 78.84 & 78.79 \\
Mistral-7B-Instruct & 90.10 & 17.89 & 66.13 & 66.13 & 55.97 & 64.94 & 58.76 & 58.76 \\
Mixtral-8x22B-Instruct & 80.20 & 58.13 & 72.87 & 72.85 & 71.82 & 65.37 & 69.81 & 69.81 \\
Qwen3-235B-A22B & 81.21 & 76.02 & 79.49 & 79.37 & 85.13 & 56.71 & 76.28 & 76.21 \\
QwQ-32B & 85.45 & 69.51 & 80.16 & 79.92 & 84.34 & 60.17 & 76.82 & 76.70 \\
DeepSeek-R1 & 87.07 & 61.79 & 78.68 & 75.60 & 79.65 & 60.61 & 73.72 & 69.81 \\
Phi-3-Medium-4K-Instruct & 85.66 & 21.14 & 64.24 & 64.22 & 36.59 & 61.90 & 44.47 & 44.40 \\
DeepSeek-V3.1 & 70.10 & 82.52 & 74.22 & 74.20 & 76.71 & 64.07 & 72.78 & 72.76 \\
DeepSeek-V3.2 & 64.44 & 90.65 & 73.14 & 73.13 & 62.43 & 69.70 & 64.69 & 63.95 \\
Kimi-K2-Thinking & 63.43 & 68.29 & 65.05 & 62.30 & 58.12 & 67.53 & 61.05 & 58.64 \\
MiniMax-M2.1 & 71.52 & 70.73 & 71.26 & 70.51 & 71.76 & 57.58 & 67.34 & 67.30 \\
MiniMax-M2.5 & 86.44 & 71.95 & 81.62 & 81.60 & 80.82 & 58.01 & 73.72 & 73.72 \\
Qwen3.5-397B-A17B & 68.89 & 84.55 & 74.09 & 74.09 & 65.75 & 77.92 & 69.54 & 69.54 \\
\bottomrule
\end{tabular}
\caption{Dataset-specific CoT-prompting performance on Task 3 (noisy belief-change detection).}
\label{tab:task3-cot}
\end{table*}

\begin{table*}[!htbp]
\centering
\small
\setlength{\tabcolsep}{4pt}
\begin{tabular}{l *{8}{S[table-format=2.2]}}
\toprule
\multirow{2}{*}{Model Name} & \multicolumn{4}{c}{\textbf{bigtom}} & \multicolumn{4}{c}{\textbf{opentom}} \\
\cmidrule(lr){2-5} \cmidrule(lr){6-9}
 & {TPR} & {TNR} & {Acc} & {F1} & {TPR} & {TNR} & {Acc} & {F1} \\
\midrule
GPT-4o & 75.56 & 23.17 & 58.16 & 57.94 & 81.21 & 47.19 & 70.62 & 70.53 \\
GPT-3.5-Turbo & 82.63 & 10.98 & 58.84 & 58.85 & 92.37 & 21.21 & 70.22 & 70.14 \\
GPT-OSS-120B & 89.90 & 15.85 & 65.32 & 65.32 & 95.50 & 43.29 & 79.25 & 79.12 \\
Llama-8B & 79.80 & 10.16 & 56.68 & 56.65 & 91.19 & 13.42 & 66.98 & 66.75 \\
Llama-405B & 66.26 & 29.27 & 53.98 & 53.75 & 85.91 & 38.53 & 71.16 & 71.17 \\
Llama-4-Maverick-17B & 70.91 & 17.07 & 53.04 & 53.06 & 87.67 & 31.17 & 70.08 & 69.99 \\
Mistral-7B-Instruct & 85.45 & 0.81 & 57.35 & 57.34 & 81.80 & 14.72 & 60.92 & 60.85 \\
Mixtral-8x22B-Instruct & 72.73 & 13.01 & 52.90 & 52.95 & 77.10 & 40.69 & 65.77 & 65.94 \\
Qwen3-235B-A22B & 69.29 & 35.77 & 58.16 & 58.14 & 80.82 & 41.13 & 68.46 & 68.49 \\
QwQ-32B & 78.38 & 20.33 & 59.11 & 59.34 & 83.95 & 38.53 & 69.81 & 69.98 \\
DeepSeek-R1 & 74.75 & 33.33 & 61.00 & 60.90 & 87.87 & 51.95 & 76.68 & 76.67 \\
Phi-3-Medium-4K-Instruct & 92.73 & 3.25 & 63.02 & 62.82 & 62.04 & 40.69 & 55.39 & 55.34 \\
DeepSeek-V3.1 & 61.62 & 46.75 & 56.68 & 56.46 & 80.23 & 51.52 & 71.29 & 71.28 \\
DeepSeek-V3.2 & 55.35 & 41.87 & 50.88 & 50.79 & 80.04 & 55.41 & 72.37 & 72.38 \\
Kimi-K2-Thinking & 59.19 & 39.02 & 52.50 & 52.38 & 81.60 & 54.11 & 73.05 & 73.12 \\
MiniMax-M2.1 & 59.84 & 29.27 & 49.66 & 49.66 & 78.04 & 45.41 & 67.93 & 68.13 \\
MiniMax-M2.5 & 77.58 & 26.02 & 60.46 & 60.44 & 92.76 & 41.56 & 76.82 & 76.78 \\
Qwen3.5-397B-A17B & 61.01 & 54.07 & 58.70 & 58.66 & 87.87 & 58.87 & 78.84 & 78.86 \\
\bottomrule
\end{tabular}
\caption{Dataset-specific CoT-prompting performance on Task 4 (noisy updated-belief identification).}
\label{tab:task4-cot}
\end{table*}

\begin{table*}[!htbp]
    \centering
    \small
    \renewcommand{\arraystretch}{1.2}
    \begin{tabular}{|p{0.95\linewidth}|}
        \hline
        \rowcolor{gray!15} \textbf{\large BigToM Generation Prompt Template} \\
        \hline
        \vspace{2pt}
        \textbf{Belief-changing cue generation:} \\
        Story: \{Context of Story Scenario from BigToM\} \{Protagonist's Desires in Story Scenario from BigToM\} \{Initial Binary Perception Cues in Story Scenario from BigToM\} \\
        Main character's belief: \{Protagonist's Belief in Story Scenario from BigToM\} \\
        Event: \{Causal Event in Story Scenario from BigToM\} \\
        Instructions: Based on the following story and event, generate perceptual cues that the main character might encounter, which MUST challenge their belief. Each cue should be creative, reasonable, and systematically explore the sensory modalities of: visual, auditory, olfactory, gustatory, tactile, proprioception (sense of body position/weight). Be descriptive and only make vivid descriptions about the perceptual cue, without describing the main character's thoughts. Avoid making explicit comparisons between the event and the main character's belief. Generate 2 perceptual cues for each sensory modality. Be super duper creative! Do not use these words: different, difference, instead of, contrasts. \\
        For each cue, write your output in a format like this example: \\
        \#\#\# Visual Cues: \\
        1. \{cue \#1\} \\
        2. \{cue \#2\} \\
        \vspace{5pt}
        \textbf{Random-event cue generation:} \\
        The same template is used, replacing \{Causal Event in Story Scenario from BigToM\} with \{Random Event in Story Scenario from BigToM\} and only generating one perceptual cue per sensory modality instead of two. \\
        \hline
    \end{tabular}
    \caption{The generation prompt template used for BigToM-based MuSeToM instances.}
    \label{tab:bigtom-generation-prompt}
\end{table*}

\begin{table*}[!htbp]
    \centering
    \small
    \renewcommand{\arraystretch}{1.2}
    \begin{tabular}{|p{0.95\linewidth}|}
        \hline
        \rowcolor{gray!15} \textbf{\large OpenToM Generation Prompt Template} \\
        \hline
        \vspace{2pt}
        \textbf{Belief-changing event generation:} \\
        Story: \{Story Scenario from OpenToM\} \\
        Main character's belief: \{Protagonist's Belief in Story Scenario from OpenToM\} \\
        Instructions: Based on the following story, generate a one-line story consisting an event that MUST NEGATE \{protagonist name\}'s initial belief and change it to: ``\{new belief\}''. Note that this story MUST also be consistent with \{observed character\}'s preference, i.e., \{observed character's preference\}. Each story should be creative, reasonable, realistic, and systematically explore the sensory modalities of: visual, auditory, olfactory, gustatory, tactile, proprioception (sense of body position/weight). Generate 2 stories for each sensory modality. Be super duper creative! Show, don't tell. Write the stories in this format: \\
        \#\#\# Visual Stories: \\
        1. \{story \#1\} \\
        2. \{story \#2\} \\
        \vspace{5pt}
        \textbf{Random-event generation:} \\
        Story: \{Story Scenario from OpenToM\} \\
        Main character's belief: \{Protagonist's Belief in Story Scenario from OpenToM\} \\
        Instructions: Based on the following story, generate a one-line story consisting a random event that occurs in the environment, which is unrelated to the object and cannot influence the main character \{protagonist name\}'s belief. Each story should be creative, reasonable, realistic, and systematically explore the sensory modalities of: visual, auditory, olfactory, gustatory, tactile, proprioception (sense of body position/weight). Generate 1 story for each sensory modality. Be super duper creative! Show, don't tell. Write the stories in this format: \\
        \#\#\# Visual Stories: \\
        \{story \#1\} \\
        \hline
    \end{tabular}
    \caption{The generation prompt templates used for OpenToM-based MuSeToM instances.}
    \label{tab:opentom-generation-prompt}
\end{table*}

\begin{table*}[!htbp]
    \centering
    \small
    \renewcommand{\arraystretch}{1.2}
    \begin{tabular}{|p{0.95\linewidth}|}
        \hline
        \rowcolor{gray!15} \textbf{\large MuSeToM Task 1 Prompt Template: Belief Change Detection} \\
        \hline
        \vspace{2pt}
        \textbf{BigToM Template:} \\
        Story: \{story\} \\
        Event: \{event\} \\
        Regarding the event, the main character \{protag\_name\} only encounters the following percept: \{percept\} \\
        Before the event, the main character's initial belief was ``\{initial\_belief\}'' \\
        Instructions: Analyze the situation and select the most appropriate answer to the following question: \\
        Did the event cause the main character \{protag\_name\} to change their belief? \\
        A. \{Yes/No - randomly shuffled\} \\
        B. \{No/Yes - randomly shuffled\} \\
        \vspace{5pt}
        \textbf{OpenToM Template:} \\
        Story: \{story\} \\
        Regarding the event, the main character \{protag\_name\} only encounters the following percept: \{percept\} \\
        Before the event, the main character's initial belief was ``\{initial\_belief\}'' \\
        Instructions: Analyze the situation and select the most appropriate answer to the following question: \\
        Did the event cause the main character \{protag\_name\} to change their belief? \\
        A. \{Yes/No - randomly shuffled\} \\
        B. \{No/Yes - randomly shuffled\} \\
        \hline
    \end{tabular}
    \caption{The zero-shot prompt template used for MuSeToM Task 1.}
    \label{tab:musetom-task1-prompt}
\end{table*}

\begin{table*}[!htbp]
    \centering
    \small
    \renewcommand{\arraystretch}{1.2}
    \begin{tabular}{|p{0.95\linewidth}|}
        \hline
        \rowcolor{gray!15} \textbf{\large MuSeToM Task 2 Prompt Template: New Belief Identification} \\
        \hline
        \vspace{2pt}
        \textbf{BigToM Template:} \\
        Story: \{story\} \\
        Event: \{event\} \\
        Regarding the event, the main character \{protag\_name\} only encounters the following percept: \{percept\} \\
        Before the event, the main character's initial belief was ``\{initial\_belief\}'' \\
        Instructions: Analyze the situation and select the most appropriate answer to the following question: \\
        After the event happened, based on the main character \{protag\_name\}'s perception of the event, which of the following statements best describes \{protag\_name\}'s new belief? \\
        Choose the most appropriate answer (choose one only): \\
        A. \{first option - randomly shuffled\} \\
        B. \{second - randomly shuffled\} \\
        C. \{third - randomly shuffled\} \\
        D. \{fourth - randomly shuffled\} \\
        \vspace{5pt}
        \textbf{OpenToM Template:} \\
        Story: \{story\} \\
        Regarding the event, the main character \{protag\_name\} only encounters the following percept: \{percept\} \\
        Before the event, the main character's initial belief was ``\{initial\_belief\}'' \\
        Instructions: Analyze the situation and select the most appropriate answer to the following question: \\
        After the event happened, based on the main character \{protag\_name\}'s perception of the event, which of the following statements best describes \{protag\_name\}'s new belief? \\
        Choose the most appropriate answer (choose one only): \\
        A. \{first option - randomly shuffled\} \\
        B. \{second - randomly shuffled\} \\
        C. \{third - randomly shuffled\} \\
        D. \{fourth - randomly shuffled\} \\
        \hline
    \end{tabular}
    \caption{The zero-shot prompt template used for MuSeToM Task 2.}
    \label{tab:musetom-task2-prompt}
\end{table*}

\begin{table*}[!htbp]
    \centering
    \small
    \renewcommand{\arraystretch}{1.2}
    \begin{tabular}{|p{0.95\linewidth}|}
        \hline
        \rowcolor{gray!15} \textbf{\large MuSeToM Task 3 Prompt Template: Noisy Belief Change Detection} \\
        \hline
        \vspace{2pt}
        \textbf{BigToM Template:} \\
        Story: \{story\} \\
        Event: \{event\} \\
        Regarding the event, the main character \{protag\_name\} only encounters the following percepts: \\
        1. \{noise/percept - randomly shuffled\} \\
        2. \{percept/noise - randomly shuffled\} \\
        Before the event, the main character's initial belief was ``\{initial\_belief\}'' \\
        Instructions: Analyze the situation and select the most appropriate answer to the following question: \\
        Did the event cause the main character \{protag\_name\} to change their belief? \\
        A. \{Yes/No - randomly shuffled\} \\
        B. \{No/Yes - randomly shuffled\} \\
        \vspace{5pt}
        \textbf{OpenToM Template:} \\
        Story: \{story\} \\
        Regarding the event, the main character \{protag\_name\} only encounters the following percepts: \\
        1. \{noise/percept - randomly shuffled\} \\
        2. \{percept/noise - randomly shuffled\} \\
        Before the event, the main character's initial belief was ``\{initial\_belief\}'' \\
        Instructions: Analyze the situation and select the most appropriate answer to the following question: \\
        Did the event cause the main character \{protag\_name\} to change their belief? \\
        A. \{Yes/No - randomly shuffled\} \\
        B. \{No/Yes - randomly shuffled\} \\
        \hline
    \end{tabular}
    \caption{The zero-shot prompt template used for MuSeToM Task 3.}
    \label{tab:musetom-task3-prompt}
\end{table*}

\begin{table*}[!htbp]
    \centering
    \small
    \renewcommand{\arraystretch}{1.2}
    \begin{tabular}{|p{0.95\linewidth}|}
        \hline
        \rowcolor{gray!15} \textbf{\large MuSeToM Task 4 Prompt Template: Noisy New Belief Identification} \\
        \hline
        \vspace{2pt}
        \textbf{BigToM Template:} \\
        Story: \{story\} \\
        Event: \{event\} \\
        Regarding the event, the main character \{protag\_name\} only encounters the following percepts: \\
        1. \{noise/percept - randomly shuffled\} \\
        2. \{percept/noise - randomly shuffled\} \\
        Before the event, the main character's initial belief was ``\{initial\_belief\}'' \\
        Instructions: Analyze the situation and select the most appropriate answer to the following question: \\
        After the event happened, based on the main character \{protag\_name\}'s perception of the event, which of the following statements best describes \{protag\_name\}'s new belief? \\
        Choose the most appropriate answer (choose one only): \\
        A. \{first option - randomly shuffled\} \\
        B. \{second - randomly shuffled\} \\
        C. \{third - randomly shuffled\} \\
        D. \{fourth - randomly shuffled\} \\
        \vspace{5pt}
        \textbf{OpenToM Template:} \\
        Story: \{story\} \\
        Regarding the event, the main character \{protag\_name\} only encounters the following percepts: \\
        1. \{noise/percept - randomly shuffled\} \\
        2. \{percept/noise - randomly shuffled\} \\
        Before the event, the main character's initial belief was ``\{initial\_belief\}'' \\
        Instructions: Analyze the situation and select the most appropriate answer to the following question: \\
        After the event happened, based on the main character \{protag\_name\}'s perception of the event, which of the following statements best describes \{protag\_name\}'s new belief? \\
        Choose the most appropriate answer (choose one only): \\
        A. \{first option - randomly shuffled\} \\
        B. \{second - randomly shuffled\} \\
        C. \{third - randomly shuffled\} \\
        D. \{fourth - randomly shuffled\} \\
        \hline
    \end{tabular}
    \caption{The zero-shot prompt template used for MuSeToM Task 4.}
    \label{tab:musetom-task4-prompt}
\end{table*}

\begin{table*}[!htbp]
\centering
\small
\renewcommand{\arraystretch}{1.15}
\begin{tabularx}{\textwidth}{p{0.22\textwidth}X}
\toprule
\textbf{Check Type} & \textbf{Prompt Template} \\
\midrule
Free generation & Give a \textbf{\texttt{\{BigToM/OpenToM\}}} test example and its answer. \\
Partial reconstruction & You are provided with Sentence 1 from the sampled \textbf{\texttt{\{BigToM/OpenToM\}}} benchmark instance. Finish Sentence 2 as it appears in the dataset. Sentence 2 must exactly match the instance in the dataset. Sentence 1: \textbf{\texttt{<Sentence1-of-Example>}} \\
\bottomrule
\end{tabularx}
\caption{Prompt templates used for potential contamination checking. Free Generation follows the dataset-name free-generation protocol of Li and Flanigan~\citep{li2024task}; Partial Reconstruction follows the partial-instance reconstruction protocol of Golchin and Surdeanu~\citep{golchin2024time}.}
\label{tab:contamination_prompt_template}
\end{table*}

\clearpage
\end{document}